\documentclass[letterpaper]{article} 
\usepackage{aaai25}  
\usepackage{times}  
\usepackage{helvet}  
\usepackage{courier}  
\usepackage[hyphens]{url}  
\usepackage{graphicx} 
\usepackage{natbib}  
\usepackage{caption} 
\usepackage{caption}
\usepackage{tikz}
\usetikzlibrary{arrows,shapes,snakes,automata,backgrounds,petri}

\usepackage{algorithm}
\usepackage{algorithmic}

\usepackage{newfloat}
\usepackage{listings}
\usepackage{xcolor}
\usepackage{booktabs} 
\usepackage{multirow} 
\usepackage{xcolor} 
\DeclareCaptionStyle{ruled}{labelfont=normalfont,labelsep=colon,strut=off} 
\floatstyle{ruled}
\newfloat{listing}{tb}{lst}{}
\floatname{listing}{Listing}
\usepackage{amsmath}
\usepackage{amsfonts}
\usepackage{graphicx}

\newcommand{\hf}{{\frac 12}}

\newcommand{\grad}{\ensuremath{\nabla}}

\newcommand{\bfA}{{\bf A}}

\newcommand{\bfD}{{\bf D}}
\newcommand{\bfE}{{\bf E}}

\newcommand{\bfH}{{\bf H}}
\newcommand{\bfI}{{\bf I}}
\newcommand{\bfJ}{{\bf J}}

\newcommand{\bfL}{{\bf L}}

\newcommand{\bfP}{{\bf P}}

\newcommand{\bfR}{{\bf R}}

\newcommand{\bfx}{{\bf x}}

\newcommand{\bfp}{{\bf p}}

\newcommand{\bfd}{{\bf d}}

\newcommand{\bff}{{\bf f}}

\newcommand{\bfz}{{\bf z}}

\newcommand{\bfphi}{{\boldsymbol \Phi}}

\newcommand{\bfepsilon}{{\boldsymbol \varepsilon}}
\newcommand{\bftheta}{{\boldsymbol \theta}}

\title{Learning Regularization for Graph Inverse Problems}
\author{
    Moshe Eliasof\textsuperscript{\rm 1},
    Md Shahriar Rahim Siddiqui\textsuperscript{\rm 2}, 
    Carola-Bibiane Sch\"onlieb\textsuperscript{\rm 1},
    Eldad Haber\textsuperscript{\rm 3}
}
\affiliations{
    \textsuperscript{\rm 1}Department of Applied Mathematics, University of Cambridge, Cambridge, United Kingdom\\
    \textsuperscript{\rm 2}Department of Physics and Astronomy, University of British Columbia, Vancouver, Canada\\
    \textsuperscript{\rm 3}Department of Earth, Ocean and Atmospheric Sciences, University of British Columbia, Vancouver, Canada\\
    me532@cam.ac.uk, shahsiddiqui@phas.ubc.ca, cbs31@cam.ac.uk, ehaber@eoas.ubc.ca

}

\usepackage{bibentry}
\begin{document}

\maketitle

\begin{abstract}
In recent years, Graph Neural Networks (GNNs) have been utilized for various applications ranging from drug discovery to network design and social networks. In many applications, it is impossible to observe some properties of the graph directly; instead, noisy and indirect measurements of these properties are available. These scenarios are coined as Graph Inverse Problems (GRIPs). In this work, we introduce a framework leveraging GNNs to solve GRIPs. The framework is based on a combination of likelihood and prior terms, which are used to find a solution that fits the data while adhering to learned prior information. Specifically, we propose to combine recent deep learning techniques that were developed for inverse problems, together with GNN architectures, to formulate and solve GRIPs. We study our approach on a number of representative problems that demonstrate the effectiveness of the framework.
\end{abstract}

%

\section{Introduction}

Graphs represent an elegant and powerful way to describe and analyze connected data, capturing the relationships and interactions between different entities in a structured manner. This mathematical framework allows for the exploration and exploitation of both the structure and the intrinsic properties of the data, enabling the discovery of hidden connections and dependencies that may not be evident by directly processing each data point. In contrast, by representing data as nodes and their relationships as edges, i.e., as graphs, we can facilitate a deeper understanding of complex systems, ranging from social networks and biological systems to communication networks, as well as financial markets.

In recent years,  graph machine learning frameworks were developed, predominantly, Graph Neural Networks \cite{scarselli2008graph, bronstein2021geometric}, which are able to model and learn complex patterns in graph data. These recent advancements make it possible to perform a wide array of applications, such as node classification \cite{kipf2016semi, defferrard2016convolutional},  community detection \cite{chen2018supervised}, drug discovery \cite{jiang2021could}, and solving combinatorial problems \cite{schuetz2022combinatorial, eliasof2024graph}. 

However, despite their high and vast utilization, in many practical scenarios, the direct observation of properties of the graph is not feasible or limited. For example, the knowledge of node labels might be partial. Instead, we often have access only to the outcomes of certain operations or processes that act upon the graph properties to be recovered. Additional examples are, in road network systems, where we might observe traffic patterns rather than the underlying network features, or, in social networks, we might see interaction patterns without having direct access to the strength of the network's relationships. This observation poses a significant research question: \textit{how can we recover or infer the original graph properties from these observed outcomes?}

Further complexity typically arises from the fact that this question is often ill-posed. This implies that given the data, there is more than one answer to the question and the solutions can be unstable with respect to even small perturbations in the data.

To address this question and its challenges, we begin by defining Graph Inverse Problems (GRIPs), highlighting a key difference between classical inverse problems that are commonly solved on structured grids 
and those on graphs. This difference directly stems from the presence of the graph, which provides additional structure. Furthermore, in many GRIP cases, there exist additional features on the graphs (throughout this paper, we will refer to these features as \textit{meta-data}) which may not be directly related to the observed data associated with the inverse problem, but can be leveraged to obtain better estimates for the solution of the inverse problem. 
Following the definition of such problems, and a number of examples, we discuss several viable approaches, from classical to neural methods, and evaluate them on several datasets and GRIPs. 

Importantly, we note that, although different GRIPs share many characteristics, existing methods to solve them, such as the seminal works in
\citet{NEURIPS2023_46ab9d96, huang2023twostage,ling2024source} are specialized for solving specific problems, as we discuss in Appendix ~\ref{sec:related_work}. In this work, we focus on extending these ideas to a general framework that can be applied to various GRIPs.  
Previous works on solving similar problems do not use any meta-data as additional features, and consider the likelihood (i.e., data fit) as a part of the network; Rather, they focus on applying various GNN architectures to solve the problem directly from the observed data, while in this work, we show the effectiveness of including meta-data in GRIPs. 
Finally, as we discuss in Appendix \ref{sec:related_work}, most GNN-based methods known to us that have been proposed for the solution of GRIPs use the same graph structure for training and prediction, where the variable part is the observed data, given as node features. This choice, while useful in many cases, can be limiting for many practical problems, where not only the observed data changes but also the underlying graph. These three key concepts distinguish our proposed GRIP framework from existing works.

\noindent \textbf{Main Contributions.}  
This paper advances the bridge between the field of inverse problems and graph machine learning by an overarching methodology for the solution of inverse problems that reside on graphs, that is, GRIP. By integrating the principles of learned regularization techniques \cite{adler2017solving, mardani2018neural}, with the flexibility of GNNs \cite{khemani2024review}, this work aims to develop a unified framework that leverages the strengths of both approaches, to solve GRIP. We demonstrate our framework on several key GRIPs, such as graph feature completion, source estimation, inverse graph transport, and the nonlinear edge recovery problem. We note that the proposed framework not only enhances the existing techniques for GRIP but also broadens the scope of GNN applications.

\section{Graph Inverse Problems}
\label{sec:graph_inverse_problems_definition}
In this section, we define and provide background on inverse problems defined on graphs, followed by examples of several key inverse problems with real-world applications.

\noindent \textbf{Notations.} Throughout this paper, 
we consider input features, that reside on a graph ${\cal G}$ defined by its set of $n$ nodes ${\cal V}$ and $m$ edges ${\cal E}$.
The features can be associated with either nodes or edges, or both, depending on the inverse problem. 

In addition, we consider an observation that is a function of the recoverable properties, and the graph structure.
To be more precise, the features are divided into {\em meta-data}, $\bff_{\rm{M}} \in {\cal F}$, and states that describe the properties, denoted by
$\bfx = [\bfx_N, \bfx_E] \in {\cal X}$. The state $\bfx_N$ is associated with the nodes and the state $\bfx_E$ is 
associated with the edge. We assume that we have observed data, i.e., measurements, $\bfd^{\rm obs} \in {\cal D}$,  on the states $\bfx$. Note that, the difference between the meta-data $\bff_{\rm{M}}$, the states, $\bfx$, and the observed data $\bfd^{\rm obs}$ is important; While $\bff_{\rm{M}}$ and $\bfx$ reside on the nodes or edges of the graph, the observed data $\bfd^{\rm obs}$, in general, reside in a different space that does not share the same domain. An example is having observed data that is related to a global measurement from the graph, e.g., average node degree.

In the context of inverse problems, the observed data is the observation from which the desired state is to be recovered, while the (optional) meta-data serves as additional information that can be used in the inverse problem solution process. Namely, we assume the following connection between the observed data $\bfd^{\rm obs}$ and the state $\bfx$:
\begin{eqnarray}
    \label{eq:measurement}
    \bfd^{\rm obs} = F(\bfx; {\cal G}) + \bfepsilon
\end{eqnarray}
where 
$F:{\cal X} \rightarrow {\cal D}$ is the forward map of the problem 
that maps the states $\bfx$ that reside on the  graph to the data space ${\cal D}$. The vector $\bfepsilon$ is some noise that, for simplicity, is assumed to be $\sigma \sim N(0, \sigma^2 \bfI)$.
While for some problems, the noise can be substantial, for many practical problems (e.g., graph segmentation), the noise can be negligent.

In the \emph{forward} problem, we are given the graph ${\cal G}$, the states $\bfx$, from which we can obtain the forward problem data, $F(\bfx, {\cal G})$. In the \emph{inverse} problem, we are given possibly noisy observations, $\bfd^{\rm obs}$, the graph ${\cal G}$, and optionally meta-data, $\bff_{\rm{M}}$. The goal is to estimate the states $\bfx$. 
In what follows, we define and discuss several key graph inverse problems. 

\begin{figure}[h]
    \centering
    \begin{tikzpicture}
    \node[shape=circle,draw=black, fill=red] (A) at (0,0) {1};
    \node[shape=circle,draw=red, fill=black!50] (B) at (0,1.5) {2};
    \node[shape=circle,draw=black, fill=red] (C) at (1.25,2) {3};
    \node[shape=circle,draw=black, fill=black!50] (D) at (1.25,0.5) {4};
    \node[shape=circle,draw=black, fill=black!50] (E) at (1.25,-1.5) {5};
    \node[shape=circle,draw=black, fill=red] (F) at (2.5,1.5) {6} ;

    \path [-] (A) edge node[left] {} (B);
    \path [-](B) edge node[left] {} (C);
    \path [-](A) edge node[left] {} (D);
    \path [-](D) edge node[left] {} (C);
    \path [-](A) edge node[right] {} (E);
    \path [-](D) edge node[left] {} (E);
    \path [-](D) edge node[right] {} (F);
    \path [-](C) edge node[right] {} (F);
    \path [-](E) edge node[right] {} (F); 

    \node[shape=circle,draw=black, fill=red] (G) at (4,0) {1};
    \node[shape=circle,draw=red, fill=black!50] (H) at (4,1.5) {2};
    \node[shape=circle,draw=black] (I) at (5.25,2) {?};
    \node[shape=circle,draw=black] (J) at (5.25,0.5) {?};
    \node[shape=circle,draw=black] (K) at (5.25,-1.5) {?};
    \node[shape=circle,draw=black] (L) at (6.5,1.5) {?} ;

    \path [-] (G) edge node[left] {} (H);
    \path [-](H) edge node[left] {} (I);
    \path [-](G) edge node[left] {} (J);
    \path [-](J) edge node[left] {} (I);
    \path [-](G) edge node[right] {} (K);
    \path [-](J) edge node[left] {} (K);
    \path [-](J) edge node[right] {} (L);
    \path [-](I) edge node[right] {} (L);
    \path [-](K) edge node[right] {} (L); 

    \draw [->,>=stealth] (2,-1.5) -- (4,-1.5);
    \node[]  at (3,-1.3) {$\bfI_n^{\bfp}$};
\end{tikzpicture}
    \caption{The forward problem of graph density completion.}
    \label{fig:gc}
\end{figure}
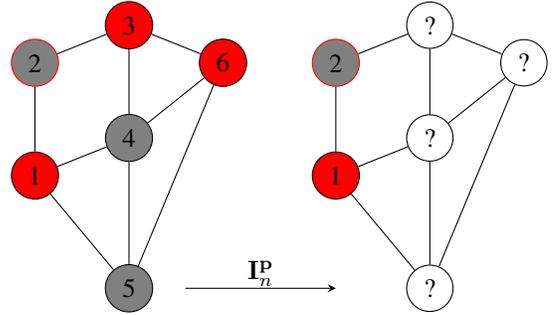

\subsection{Property Completion (Figure \ref{fig:gc})} 

Consider the case where the forward problem reads
\begin{equation}
    \label{eq:maskingOperator}
    F(\bfx, {\cal G}) = \bfI_n^{\bfP} \bfx, 
\end{equation}
where $\bfI_n^{\bfp} \in \mathbb{R}^{p\times n}$ is a subset of $p$ rows from  an $n
\times n$ identity matrix.  In this case, we need to reconstruct the graph properties from the partial, potentially noisy data $\bfd^{\rm obs}$. This problem is commonly solved in the GNN literature (see e.g., \cite{bronstein2017geometric}). We note that, in the absence of noise, i.e. $\epsilon=0$, the inverse problem defined by the operator in Equation \eqref{eq:maskingOperator} coincides with the typical settings of semi-supervised node classification \cite{kipf2016semi}.
An illustration of the problem is provided in Figure~\ref{fig:gc}.

\begin{figure}[h]
    \centering
    \begin{tikzpicture}
    \node[shape=circle,draw=black] (A) at (0,0) {1};
    \node[shape=circle,draw=red, fill=red] (B) at (0,1.5) {2};
    \node[shape=circle,draw=black] (C) at (1.25,2) {3};
    \node[shape=circle,draw=black] (D) at (1.25,0.5) {4};
    \node[shape=circle,draw=black] (E) at (1.25,-1.5) {5};
    \node[shape=circle,draw=black] (F) at (2.5,1.5) {6} ;

    \path [-] (A) edge node[left] {} (B);
    \path [-](B) edge node[left] {} (C);
    \path [-](A) edge node[left] {} (D);
    \path [-](D) edge node[left] {} (C);
    \path [-](A) edge node[right] {} (E);
    \path [-](D) edge node[left] {} (E);
    \path [-](D) edge node[right] {} (F);
    \path [-](C) edge node[right] {} (F);
    \path [-](E) edge node[right] {} (F); 

    \node[shape=circle,draw=black, fill=red!20] (G) at (4,0) {1};
    \node[shape=circle,draw=red, fill=red!70] (H) at (4,1.5) {2};
    \node[shape=circle,draw=black, fill=red!10] (I) at (5.25,2) {3};
    \node[shape=circle,draw=black, fill=red!50] (J) at (5.25,0.5) {4};
    \node[shape=circle,draw=black, fill=red!10] (K) at (5.25,-1.5) {5};
    \node[shape=circle,draw=black, fill=red!40] (L) at (6.5,1.5) {6} ;

    \path [-] (G) edge node[left] {} (H);
    \path [-](H) edge node[left] {} (I);
    \path [-](G) edge node[left] {} (J);
    \path [-](J) edge node[left] {} (I);
    \path [-](G) edge node[right] {} (K);
    \path [-](J) edge node[left] {} (K);
    \path [-](J) edge node[right] {} (L);
    \path [-](I) edge node[right] {} (L);
    \path [-](K) edge node[right] {} (L); 

    \draw [->,>=stealth] (2,-1.5) -- (4,-1.5);
    \node[]  at (3,-1.3) {$\bfP^k$};
\end{tikzpicture}
    \caption{A source graph on the left is diffused over time, with transition matrix $\bfP^k$ obtaining a target graph on the right. The goal of the inverse problem is to identify the source given the target.}
    \label{fig:sl}
\end{figure}
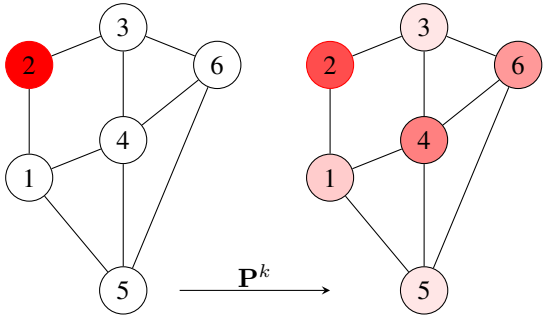

\subsection{Inverse Source Estimation (Figure \ref{fig:sl})} 
A second popular problem is inverse source estimation, sometimes referred to as the source localization problem \cite{huang2023twostage}. This problem occurs by modeling the spread of information on a graph.
Specifically, we consider the case where the forward problem reads
\begin{equation}
    \label{eq:sourceProblem}
    F(\bfx, {\cal G}) = \bfP^k \bfx^0,
\end{equation}
where $\bfP$ is a Markov transition matrix that spreads the information from nodes to their neighbors. 
In the canonical setting, $k$ represents some time frame where the information spreads from one node to its neighbors. If we observe the system after $k$ time frames, we obtain \eqref{eq:sourceProblem}. 
A popular instance of this problem is the case where $\bfP$ is the degree normalized adjacency matrix, i.e., $\bfP = \bfD^{-1}\bfA$, where $\bfD$ is the node degree matrix, and $\bfA$ is the binary adjacency matrix. The goal is then to find the source $\bfx^0$, as illustrated in Figure~\ref{fig:sl}.

At this point, it is important to note that for the source estimation problem, the problem becomes harder as $k$ becomes larger because an increase in $k$ exponentially damps the information in $\bfx$ that is associated with eigenvalues that are smaller than $1$ \cite{gvl}.

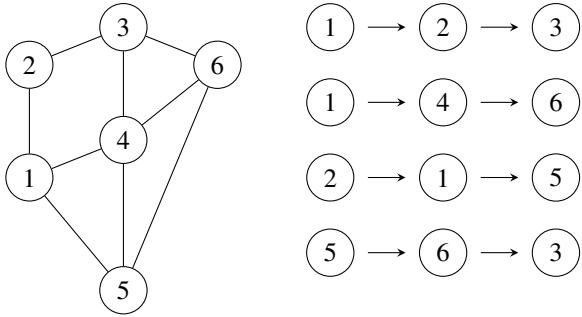
\begin{figure}
    \centering
    \begin{tikzpicture}
    \node[shape=circle,draw=black] (A) at (0,0) {1};
    \node[shape=circle,draw=black] (B) at (0,1.5) {2};
    \node[shape=circle,draw=black] (C) at (1.25,2) {3};
    \node[shape=circle,draw=black] (D) at (1.25,0.5) {4};
    \node[shape=circle,draw=black] (E) at (1.25,-1.5) {5};
    \node[shape=circle,draw=black] (F) at (2.5,1.5) {6} ;

    \path [-] (A) edge node[left] {} (B);
    \path [-](B) edge node[left] {} (C);
    \path [-](A) edge node[left] {} (D);
    \path [-](D) edge node[left] {} (C);
    \path [-](A) edge node[right] {} (E);
    \path [-](D) edge node[left] {} (E);
    \path [-](D) edge node[right] {} (F);
    \path [-](C) edge node[right] {} (F);
    \path [-](E) edge node[right] {} (F); 

    \node[shape=circle,draw=black] at (4,2)    {1};
    \draw [->,>=stealth] (4.5,2) -- (5,2);
    \node[shape=circle,draw=black] at (5.5, 2) {2};
    \draw [->,>=stealth] (6,2) -- (6.5,2);
    \node[shape=circle,draw=black] at (7, 2) {3};

    \node[shape=circle,draw=black] at (4,1)    {1};
    \draw [->,>=stealth] (4.5,1) -- (5,1);
    \node[shape=circle,draw=black] at (5.5, 1) {4};
    \draw [->,>=stealth] (6,1) -- (6.5,1);
    \node[shape=circle,draw=black] at (7, 1) {6};

    \node[shape=circle,draw=black] at (4,0)    {2};
    \draw [->,>=stealth] (4.5,0) -- (5,0);
    \node[shape=circle,draw=black] at (5.5, 0) {1};
    \draw [->,>=stealth] (6,0) -- (6.5,0);
    \node[shape=circle,draw=black] at (7, 0) {5};

    \node[shape=circle,draw=black] at (4,-1)    {5};
    \draw [->,>=stealth] (4.5,-1) -- (5,-1);
    \node[shape=circle,draw=black] at (5.5, -1) {6};
    \draw [->,>=stealth] (6,-1) -- (6.5,-1);
    \node[shape=circle,draw=black] at (7, -1) {3};

\end{tikzpicture}
    \caption{The graph transport problem. The four sampled paths of length $3$ on the right are obtained from the graph on the left. The data is the sum of the node properties.  }
    \label{fig:gtomo}
\end{figure}

\subsection{Inverse Graph Transport (Figure \ref{fig:gtomo})} 
We consider the inverse graph transport, a problem where the measurement
is obtained through an averaging of the state along a walk on the graph.
Specifically, we consider the case where the observed data can be expressed as
\begin{equation}
    \label{eq:graphTrsnportProblem}
    F(\bfx, {\cal G})_k = \sum_{j=0}^{L-1} \bfx_{\Gamma_k^j} \quad k=1,\ldots, K
\end{equation}
Here, we are given a set of K paths $\{\Gamma_{k} \}_{k=1}^{K}$, of length $L$, such that the path $\Gamma_k$ is a vector of length $L$ whose entries are the node indices in which the path traverses through. In Equation \eqref{eq:graphTrsnportProblem} we use the notation $\Gamma_k^j$ to obtain the $j$-th node index on the path $\Gamma_k$, in order to sum over the node features that are in the path. We note that the graph inverse transport problem is similar to the ray tomography problem in a continuous setting \cite{natterer2001mathematics}. The goal is to recover the original state $\bfx$ given observations on their partial sums. 

For this problem, it is important to note that the data does not have an obvious graph structure. In fact, the data can be much larger than the graph. Some nodes can be sampled multiple times and some very few or not at all.

\noindent \textbf{Remark.} The problems defined in Equations \eqref{eq:maskingOperator}--\eqref{eq:graphTrsnportProblem} are linear. Next, we show an example of a \emph{nonlinear} problem, which is considered more complex \cite{ijerph18094432}. 

\begin{figure}[h]
    \centering
    \begin{tikzpicture}
    \node[shape=circle,draw=black] (A) at (0,0) {1};
    \node[shape=circle,draw=red, fill=red] (B) at (0,1.5) {2};
    \node[shape=circle,draw=black] (C) at (1.25,2) {3};
    \node[shape=circle,draw=black] (D) at (1.25,0.5) {4};
    \node[shape=circle,draw=black] (E) at (1.25,-1.5) {5};
    \node[shape=circle,draw=black] (F) at (2.5,1.5) {6} ;

    \path [-] (A) edge[draw=blue, very thick,opacity=0.45] node[left] {} (B);
    \path [-](B) edge[draw=blue, very thick, opacity=0.25] node[left] {} (C);
    \path [-](A) edge[draw=blue, very thick,  opacity=0.15] node[left] {} (D);
    \path [-](D) edge[draw=blue, very thick, opacity=0.35] node[left] {} (C);
    \path [-](A) edge[draw=blue, very thick] node[right] {} (E);
    \path [-](D) edge[draw=blue, very thick, opacity=0.25] node[left] {} (E);
    \path [-](D) edge[draw=blue, very thick] node[right, opacity=0.25] {} (F);
    \path [-](C) edge[draw=blue, very thick, opacity=0.15] node[right] {} (F);
    \path [-](E) edge[draw=, very thick, opacity=0.1] node[right] {} (F); 

    \node[shape=circle,draw=black, fill=red!20] (G) at (4,0) {1};
    \node[shape=circle,draw=red, fill=red!70] (H) at (4,1.5) {2};
    \node[shape=circle,draw=black, fill=red!10] (I) at (5.25,2) {3};
    \node[shape=circle,draw=black, fill=red!50] (J) at (5.25,0.5) {4};
    \node[shape=circle,draw=black, fill=red!10] (K) at (5.25,-1.5) {5};
    \node[shape=circle,draw=black, fill=red!40] (L) at (6.5,1.5) {6} ;

    \path [-] (G) edge[draw=blue, very thick,opacity=0.45] node[left] {} (H);
    \path [-](H) edge[draw=blue, very thick, opacity=0.25] node[left] {} (I);
    \path [-](G) edge[draw=blue, very thick,  opacity=0.15] node[left] {} (J);
    \path [-](J) edge[draw=blue, very thick, opacity=0.35] node[left] {} (I);
    \path [-](G) edge[draw=blue, very thick] node[right] {} (K);
    \path [-](J) edge[draw=blue, very thick, opacity=0.25] node[left] {} (K);
    \path [-](J) edge[draw=blue, very thick] node[right] {} (L);
    \path [-](I) edge[draw=blue, very thick, opacity=0.15] node[right] {} (L);
    \path [-](K) edge[draw=, very thick, opacity=0.1] node[right] {} (L); 

    \draw [->,>=stealth] (2,-1.5) -- (4,-1.5);
    \node[]  at (3,-1.3) {\textcolor{blue}{$\bfP^k$}};
\end{tikzpicture}
    \caption{A {\bf known} source graph on the left is diffused over time, with an {\bf unknown} transition matrix $\bfP^k$ determined by the edge weights, obtaining the target graph on the right. The goal of the inverse problem is to identify the edge weights given the source and the target.}
    \label{fig:er}
\end{figure}
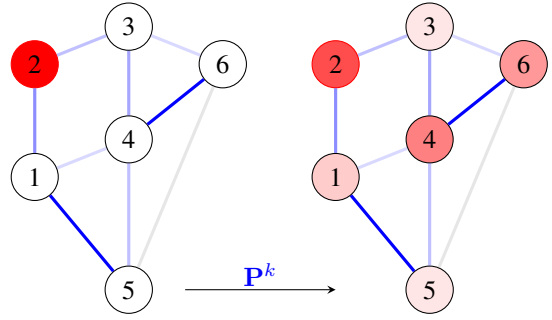

\subsection{\bf Edge Property Recovery (Figure \ref{fig:er})}

\noindent Consider the case of source estimation in Equation \eqref{eq:sourceProblem}, but with {\bf known} source and {\bf unknown} edge properties where 
\begin{eqnarray}
    \label{den}
    \bfx_{N}^{(k+1)} = \bfP(\bfx_E) \bfx_{N}^{(k)} \quad k=0,\ldots, K-1.
\end{eqnarray}
Here $\bfP(\bfx_E)$ is a Markov transition matrix that depends on the property of the edge. Assuming that we have the history $\bfd^{\rm obs} = [\bfx^{(0)}_N, \ldots, \bfx^{(K)}_N] + \bfepsilon$,  the goal is to find the edge property, that is, the edge states, $\bfx_E$ from the data.

Note that this problem is a nonlinear extension
of the source estimation problem. Here, rather than assuming that the source $\bfx_N$ is the unknown state to be recovered, we have the matrix $\bfP$, which encodes edge properties (e.g., edge weights) as the unknown states to be recovered.
\\
For the forward problems presented in this section, their inverse problems are assumed to be ill-posed. That is, the operator $F$ in Equation \eqref{eq:measurement} has a large, non-trivial null space or, such that the singular values of its Jacobian quickly decay to $0$. This implies that  a small perturbation in the data can result in a large 
perturbation in the estimated solution (see, e.g., \cite{hansen}). 
Therefore, to obtain an estimate
of the solution, some regularization is required. Such regularization can vary from classical techniques like Tikhonov regularization \cite{Tikhonov1950} (similar to weight-decay in neural networks), smoothness-based regularization \cite{NIPS2009_68ce199e}, total-variation \cite{RudinOsherFatemi92}, as well as learnable neural regularizers \cite{adler2017solving}.
In previous works on graph inverse problem \cite{huang2023twostage}, such a regularization was achieved implicitly by learning a map, $F^{\dag}$, from the data $\bfd^{\rm obs}$ to the solution $\bfx$. Nonetheless, as has been shown for other, non-graph inverse problems \cite{adler2017solving, eliasof2023drip}, even when the data fit is a part of the training, at inference, such a map can lead to reconstructions that do not fit the observed data 
 -- which can lead to non-feasible solutions. Mathematically, these findings mean that the residual between the predicted solution and the observed data, defined as:
$$ \|F (F^{\dag}(\bfd^{\rm obs}) -  \bfd^{\rm obs}\|$$
may not be small and, in some cases, diverge. Therefore, 
in Section \ref{sec:generalInverseProblemsSolution}, we explore methodologies that blend graph-based regularization with the forward problem, yielding algorithms that adhere both to a priori information (that is, the prior learned in $F^{\dagger}$) and the given observed data $\bfd^{\rm{obs}}$.

\section{Methodologies for Inverse Problems}
\label{sec:generalInverseProblemsSolution}
In this section, we develop the ideas presented in the paper. We start by reviewing the basic concepts that are used for the solution of inverse problems and then explain how they can be adopted and modified to work with the graph structure and how meta-data can be used. Later, in Section \ref{sec:graphRegularization_actual_methods}, we prescribe specific methods to solve GRIP.

\noindent \textbf{Goal.} The objective in solving Inverse Problems is to estimate a solution $\widehat \bfx$
that (i) fits the data in Equation \eqref{eq:measurement} to some
prescribed error, and (ii) adheres to a-priori information given by a regularizer.
Two approaches to solving such problems are commonly used.
The \emph{first}, is based on a variational method and the \emph{second} is to embed the solution with an (inverse) scale-space dynamical system. We now shortly review both techniques and their behavior.

\noindent \textbf{Variational Techniques.}
\label{sec:variationalMethods}
Variational methods formulate inverse problems as an optimization task, seeking to minimize a functional that balances fidelity to the observed data with the imposition of prior knowledge or regularization (see, e.g., \cite{engl1996regularization,CalvettiSomersalo2005,Tenorio2011}).
They have a strong relation to the Maximum Aposteriori Estimate (MAP) when considering a Bayesian approach \cite{taran}.
In the context of inverse problems, the objective functional includes a data fitting term, which ensures that the solution is consistent with the observed data, and a regularization term, which incorporates prior assumptions about the solution, such as smoothness, sparsity, or other structural properties. 
Here, we consider regularization techniques
where the estimated  $\widehat \bfx$ is obtained by solving a regularized data fitting problem:
\begin{subequations}    
\begin{eqnarray}
\label{eq:GeneralVariationalScheme}
  && \widehat \bfz = {\rm arg}{\min}_{\bfz}\ \hf \|F(\bfE \bfz) - \bfd^{\rm obs}\|^2 + \alpha {\bfR}(\bfz, \bftheta ), \\  && \widehat \bfx = \bfE \widehat \bfz.
\end{eqnarray}
\end{subequations}

Here $\bfE$ is a learnable embedding function, 
and ${\bfR}(\bfz, \bftheta )$ is a regularization functional that depends on learnable parameters $\bftheta$, and $\alpha$ is a non-negative coefficient that balances the terms in Equation \eqref{eq:GeneralVariationalScheme}.
The 'art' of obtaining meaningful solutions is transformed into the learning of an appropriate regularization functional ${\bfR}$ and an embedding $\bfE$.

\noindent \textbf{Inverse Scale-Space Methods.}
\label{sec:InverseScaleSpace}
One of the difficulties in solving the optimization proposed in Variational Methods, as in Equation \eqref{eq:GeneralVariationalScheme}, is the appropriate choice of the relative weight $\alpha$, between the data fitting and regularization terms \cite{nagyHansenBook}. Inverse scale-space methods \cite{burger2006} propose an alternative approach by casting the optimization problem to a dynamical system that starts with a coarse approximation of the solution and progressively incorporates finer details.
By initially focusing on the low-frequency details and gradually refining the solution to encode high-frequency details, inverse scale-space techniques can achieve a balance between fidelity to the data and the incorporation of prior knowledge. The choice of the regularization parameter is replaced by the choice of the stopping time, which is often more straightforward.

In practice, an inverse scale-space approach is obtained by
integrating an ordinary differential equation 
of the form
\begin{equation}
\footnotesize
  {\frac {d\bfz}{dt}} = \bfE(t)^{\top}\bfJ(t)^{\top}\left( \bfd^{\rm obs} - F(\bfE(t) \bfz) \right) - s(\bfz, \bftheta(t)),
\label{eq:GeneralInverseScaleSpaceScheme}
\end{equation}
from time $0$ to time $T$, with $\bfz(0) = \bfz_0$, which is determined by the level of fitting the data.
Here, $\bfE(t)$ is a time-dependent embedding, $\bfJ(t)$ is the Jacobian of $F$ with respect to its argument, and $s$ is a function referred to as the score. 
That is, in order to use a scale space approach for the solution of the problem, one is required to choose time-dependent embedding and a score function, $s$.
The flexibility of a time-dependent embedding $\bfE$ and regularization $s$ was shown to overcome issues such as local minima \cite{eliasof2024over}. 
We note that, while Inverse scale-space methods offer more flexibility, they have less theoretical understanding compared with variational methods.
Additionally, scale space methods can be reduced to solving the problem in Equation  \eqref{eq:GeneralVariationalScheme} using gradient descent, if we choose $s = \grad \bfR$ 
and $\bfE$ to be time-invariant; however, typically, in Scale Space methods, $s$ and $\bfE$ are time-dependent.

\section{A Framework for Solving GRIP}
\label{sec:graphRegularization_actual_methods}
In Section \ref{sec:variationalMethods}, we discussed generic regularization techniques that have been used to solve general inverse problems. In this Section, we discuss how such regularization techniques can be adopted to solve GRIP. In particular, we identify two key differences between inverse problems on a \emph{structured} grid, and inverse problems on \emph{graphs}, as discussed below: 

\noindent \textbf{(i) Graph-Informed Regularizers.}
Because the problem is defined on a graph, it is possible and desirable to leverage the graph structure to obtain additional insights on the support of the problem. Regularization techniques can incorporate the connectivity and topology of the graph, enabling more informed and precise recovery of the property $\bfx$. For example, smoothness constraints can be imposed, effectively promoting solutions where neighboring nodes have similar values, which is particularly useful in applications like homophilic node classification \cite{Zhou2004}. 

\noindent \textbf{(ii) Utilization of Meta-Data in Graphs.}
While the observed data 
$\bfd^{\rm obs}$, defined in Equation \eqref{eq:measurement}, 
  depends solely on states $\bfx$, in many cases, each node or edge in the graph is associated with additional meta-data 
$\bff_{\rm{M}}$. This meta-data, which may include attributes such as node positions, edge weights, or additional features, is not required to compute 
$\bfd^{\rm obs}$, 
  but it provides valuable supplementary information that can enhance the regularization process.
  For example, in applications related to 3D Computer Vision, where point clouds and graphs are processed, it is common to have multiple types of features, such as $xyz$ coordinates, point normals, node-wise labels, as well as a global label of the graph. Additionally, we note that in the absence of observed data $\bfd^{\rm{obs}}$, the proposed framework reduces to standard GNN architectures, as discussed below. 
Overall, this two-fold consideration of the graph structure and meta-data enables a more holistic approach to solving GRIP, as we now discuss.

\subsection{Classical Graph Regularization}
\label{sec:classical} 
We now discuss two classical regularization techniques that can be used for the solution of GRIP.

\noindent \textbf{Laplacian Regularization.} A classical regularizer that has been used extensively, especially in the case of graph completion \cite{NIPS2009_68ce199e}  is based on the graph Laplacian $\bfL = \bfD - \bfA$. 
Given the graph Laplacian $\bfL$, the Laplacian smoothness regularizer is defined as
\begin{equation}
    \label{eq:LapReg}
    {\bfR}(\bfx) = \hf \bfx^{\top} \bfL \bfx.
\end{equation}
To solve the optimization problem in Equation \eqref{eq:GeneralVariationalScheme}, where $\bfE=\bfI$, it is common \cite{NIPS2009_68ce199e} to use a scaled gradient descent method that reads:
\begin{equation}
    \label{eq:scaledgd}
    \bfx_{k+1} = \bfx_k - \bfH^{-1}\left( \bfJ_k^{\top}(F(\bfx_k) - \bfd^{\rm obs}) + \alpha \bfL \bfx_k \right). 
\end{equation}
Here, $\bfH$ is a symmetric positive-definite matrix, chosen to accelerate the convergence of the method, and $\bfJ_k$ is the Jacobian of $F$ at the $k$-th step. The choice $\bfH^{-1} = \mu \bfI$ yields the gradient descent.
Alternatively, choosing $\bfH^{-1} = \mu \bfL^{-1}, \ \mu > 0$ and $\alpha=0$ we obtain the scale space iteration:
\begin{equation}
    \label{eq:scale_sp}
    \bfx_{k+1} = \bfx_k - \mu \bfL^{\dag} \bfJ_k^{\top}(F(\bfx_k) - \bfd^{\rm obs}).
\end{equation}
It was shown in \citet{calvetti2005invertible}
that early termination of the iteration in Equation \eqref{eq:scale_sp}  gives similar
results to tuning the regularization parameter $\alpha$ in Equation \eqref{eq:scaledgd}.
Thus, it is possible to use Equation \eqref{eq:scaledgd} and tune $\alpha$ as a hyper-parameter, or  Equation \eqref{eq:scale_sp} and tune the number of iterations as a hyper-parameter and obtain similar results.

\noindent \textbf{Tikhonov Regularization.} The Tikhonov Regularization technique \cite{Tikhonov1977}, which promotes solutions with low norms, is defined 
by replacing the Laplacian $\bfL$ in Equation \eqref{eq:LapReg} by the identity matrix $\bfI$.

\subsection{Neural Graph Regularization}
We discuss neural approaches for learning regularization for GRIPs. All methods include terms that are dependent on the graph $\cal G$, and are implemented using a GNN. We provide specific details on the GNN architecture in Appendix \ref{appendix:architecture}.

\noindent \textbf{Variational Methods.} 
We consider the use of variational regularization, discussed in Section \ref{sec:variationalMethods}, where the key idea is to \emph{learn} the regularization term using a GNN. In particular, as previously discussed, when working with graph data, there often exist additional, meta-data, $\bff_{M}$ for each node.  
Thus, we modify a recent regularization operator, proposed in  \cite{eliasof2023drip} to include meta-data, such that this regularization reads:
\begin{eqnarray}
\label{eq:drip}
&{\bfR}(\bfx_{1}, \ldots, \bfx_{L}, \bff_{M}, \{\bftheta{_l}\}_{l=1}^L) \\ &= \underbrace{\sum_{l=1}^{L-1} \hf \|\bfx_{l+1} - \bfx_{l} \|^2 
\nonumber}_{Kinetic}  \ +  \ \underbrace{\sum_{l=1}^{L-1}\bfphi(\bfx_{l}, \bff_{M}, {\cal G}, \bftheta_l)}_{Potential}.  
\end{eqnarray}
Here, the solution $\bfx =\bfx_{L}$ is embedded in the hidden
states 
$\bfx_{1},\ldots,\bfx_{L}$  and $\hat{\bfx} = \bfx_{L}$ is the predicted solution.

As shown in Equation \eqref{eq:drip}, the regularization employs a kinetic energy term and a potential energy term. 

We now show the effect of the meta-data $\bff_{M}$ in this regularizer. To this end, consider the minimization problem in Equation \eqref{eq:GeneralVariationalScheme}, when there is no observed data $\bfd^{\rm{obs}}$. That is, we only minimize the regularization term. Differentiating Equation \eqref{eq:drip} with respect to $\bfx$ yields the leapfrog scheme:
\begin{eqnarray}
\label{hyperGNN}
  \bfx_{k+1} = 2 \bfx_{k} - \bfx_{k-1} + \grad_{\bfx_k} \bfphi(\bfx_k, \bff_{M}, {\cal G}, \bftheta_k).
\end{eqnarray}
Equation \eqref{hyperGNN} is a residual GNN with two skip connections, also referred to as a neural network with hyperbolic dynamics \cite{ruthotto2019deep}, also studied for GNNs in \citet{eliasof2021pde, rusch2022graph}. Thus, even without data $\bfd^{\rm{obs}}$, if the meta-data $\bff_{M}$ has sufficient information about the final state $\bfx$, this regularization can be sufficient to recover it.
Having additional observed data on the final state $\bfx$ can further improve the recovery. 
In our experiments, we refer to this network as \emph{Var-GNN}.

\noindent \textbf{Inverse Scale-Space.} 
The second technique we experiment with is a modification of a learned inverse scale-space technique proposed in \cite{eliasof2024over} for solving inverse problems on a structured grid. We adapt it to operate on graph problems. 
The dynamical system in Equation \eqref{eq:GeneralInverseScaleSpaceScheme} is simply discretized in two steps, obtaining:
\begin{subequations}    
\begin{eqnarray}
 && \bfz_{k+\hf} = \bfz_k + \mu \bfE_k^{\top} \bfJ_k^{\top} \left( \bfd^{\rm obs} - F(\bfE_k \bfz_k, {\cal G}) \right) \\
 \label{eq:dsd}
 && \bfz_{k+1} = \bfz_{k+\hf} - s(\bfz_{k+\hf}, \bff_{M}, {\cal G}, \bftheta_k, t_k). 
 \label{eq:dss}
 \end{eqnarray}
 \end{subequations}
 For the embedding, $\bfE_k$, we use a linear layer and a standard multilayer GNN for $s$ with time embedding. We refer to this method as \emph{ISS-GNN}.
Note that, similarly to Var-GNN, in the absence of data $\bfd^{\rm{obs}}$, the effective network reduces to Equation \eqref{eq:dsd}, which is a residual GNN. This implies that ISS-GNN can utilize the meta-data $\bff_{\rm{M}}$ to potentially recover the state $\bfx$, similar to standard uses of GNNs.

\noindent \textbf{Proximal Methods.} 
Another approach for solving inverse problems that blends scale-space and variational methods was proposed in \cite{adler2017solving, mardani2018neural}. Here, we modify the technique for graph learning.

In Proximal methods, the concept is to use proximal gradient descent as a numerical method for the solution of the optimization problem in Equation \eqref{eq:GeneralVariationalScheme}. Such iteration reads:
\begin{subequations}
\label{eq:proximal}
\begin{eqnarray}
    \label{eq:proxa}
    &\bfx_{k+\hf} = \bfx_k - \mu \bfJ_k^{\top} (F(\bfx_k) - \bfd^{\rm obs}), \\
    \label{eq:proxb}
    &\bfx_{k+1} = {\rm argmin}_{\bfx} \hf \|\bfx - \bfx_{k+\hf}\|^2 + \bfR(\bfx, \bff_{M}, \bftheta).
\end{eqnarray}
\end{subequations}
This iteration can be computationally expensive 
due to the cost of solving the optimization problem in Equation \eqref{eq:proximal}. Therefore, \citet{adler2017solving} and  \citet{mardani2018neural} propose to learn the solution of the optimization problem in Equation \eqref{eq:proximal} with a network $s$ that is:
\begin{eqnarray}
\label{eq:prox}
\nonumber
s(\bfx_{k+\hf}, \bff_{M}, {\cal G}, \bftheta)=     {\rm argmin}_{\bfx} (\hf \|\bfx - \bfx_{k+\hf}\|^2 + \\ {\bfR}(\bfx, \cal G, \bff_{M}, \bftheta) ). 
\end{eqnarray}
This approach leads to a two-stage network, where in the first step, a gradient step is applied  (Equation \eqref{eq:proxa}), which aims to fit the data,
and in the second step, a proximal step is applied (Equation \eqref{eq:prox}).
In \citet{adler2017solving}, the iteration in Equation \eqref{eq:prox} shares the same parameters $\bftheta$ for every proximal step. 
 In \citet{mardani2018neural}, the iteration is unrolled, to learn a parameter $\bftheta_k$ 
per-iteration. This choice unties the link to variational methods but increases the flexibility of the method. In our experiments we follow \citet{mardani2018neural}, and modify the method proposed there to operate on graphs. We call this network \emph{Prox-GNN}, which unites the two steps in Equation \eqref{eq:proximal} as follows:
\begin{eqnarray}
    \label{eq:pgd}
    \bfx_{k+1} = s\left(\bfx_k - \mu \bfJ_k^{\top} (F(\bfx_k) - \bfd^{\rm obs}), \bff_{M}, {\cal G}, \bftheta_k\right),
\end{eqnarray}
where $s(\cdot)$ is a GNN that uses observed data $\bfd^{\rm{obs}}$ and the meta-data to solve the problem. 
It is interesting to observe that, in the absence of observed data, we obtain a standard feed-forward GNN that uses the meta-data to predict $\bfx$.

\section{Experiments}
We evaluate the five models discussed in Section \ref{sec:graphRegularization_actual_methods} on the diverse set of GRIPs discussed in Section \ref{sec:graph_inverse_problems_definition}. We utilize several datasets to demonstrate GRIPs, ranging from synthetic graphs (CLUSTER) to geometric datasets (ShapeNet), natural images (CIFAR10), road traffic (METR-LA), and spread of disease (Chickenpox-Hungary, shortened to CPOX). We provide full details on the datasets and our motivation for using them to demonstrate different problems, in Appendix \ref{appendix:datasets}.  
We also provide a comprehensive description of the training,  evaluation, and hyperparameter selection procedures for each problem in Appendix \ref{appendix:optimization_details}, hyperparameter sensitivity results in \ref{appendix:additional_results}, and a discussion of the complexity and runtimes of the methods, in Appendix \ref{appendix:complexity}. Our code is implemented in Pytorch \cite{paszke2017automatic}, and is available at \url{https://github.com/nafi007/GraphInverseProblems}.

\noindent \textbf{Research Questions.} In our experiments, we seek to address the following questions: (i) Can the framework proposed in this paper be applied to various GRIPs?  (ii) Do neural methods consistently perform better than classical methods, and to what extent? (iii) What is the influence of using meta-data on downstream performance?

\subsection{Property Completion}
\begin{figure*}
    \centering
    \includegraphics[trim={15cm 0 15cm 0},clip,width=0.136\linewidth]{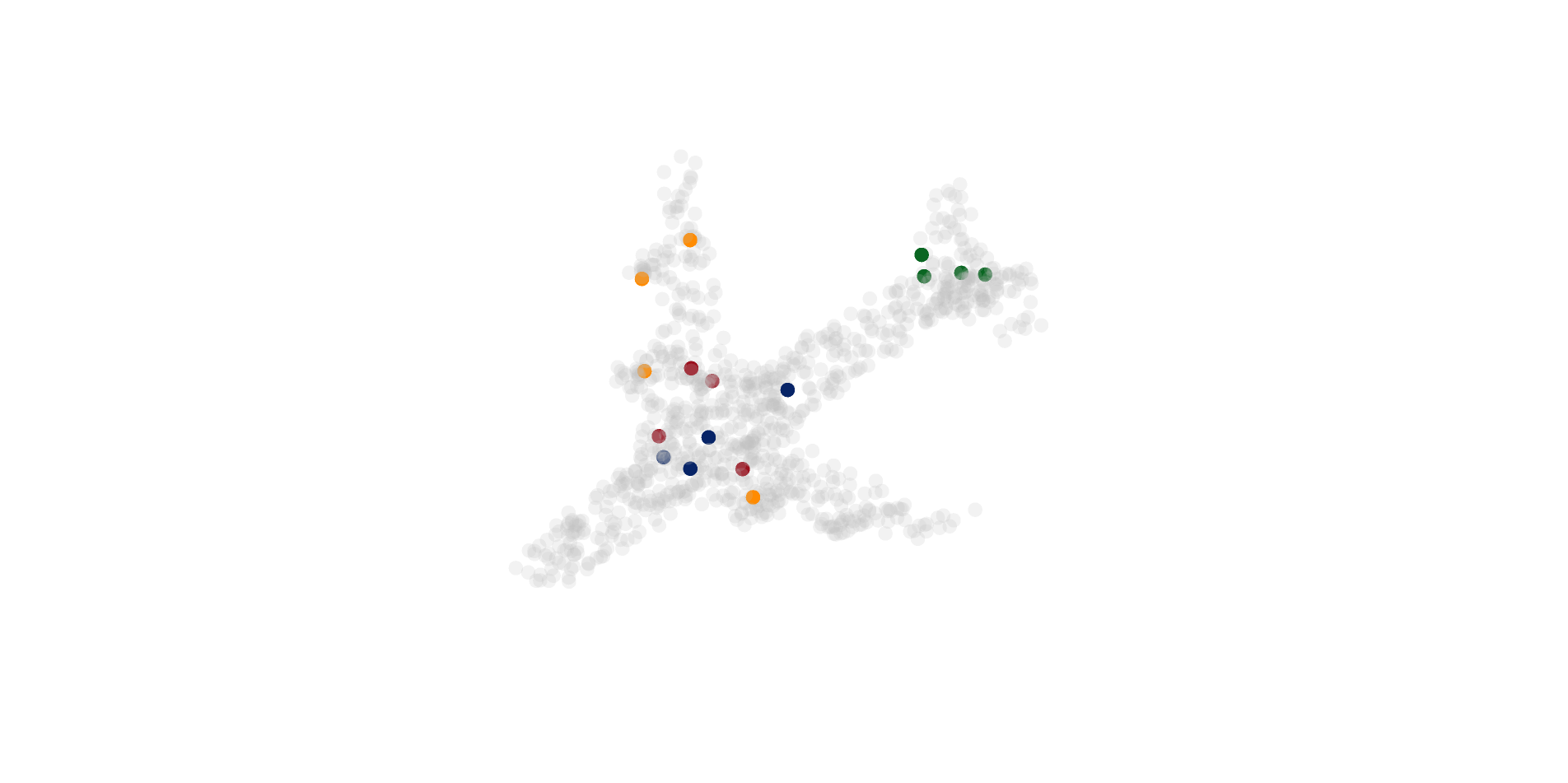}
    \includegraphics[trim={15cm 0 15cm 0},clip,width=0.136\linewidth]{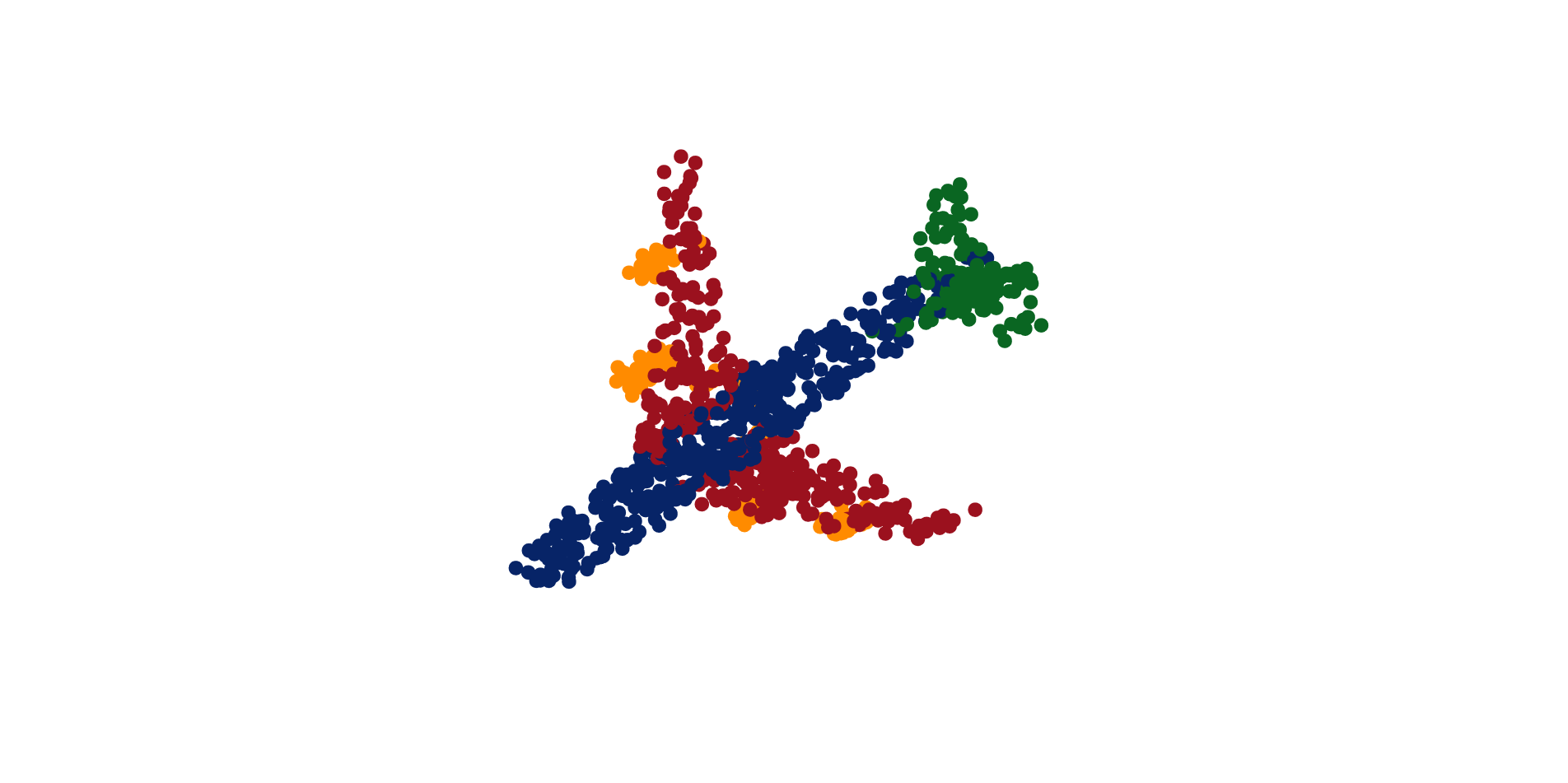}
    \includegraphics[trim={15cm 0 15cm 0},clip,width=0.136\linewidth]{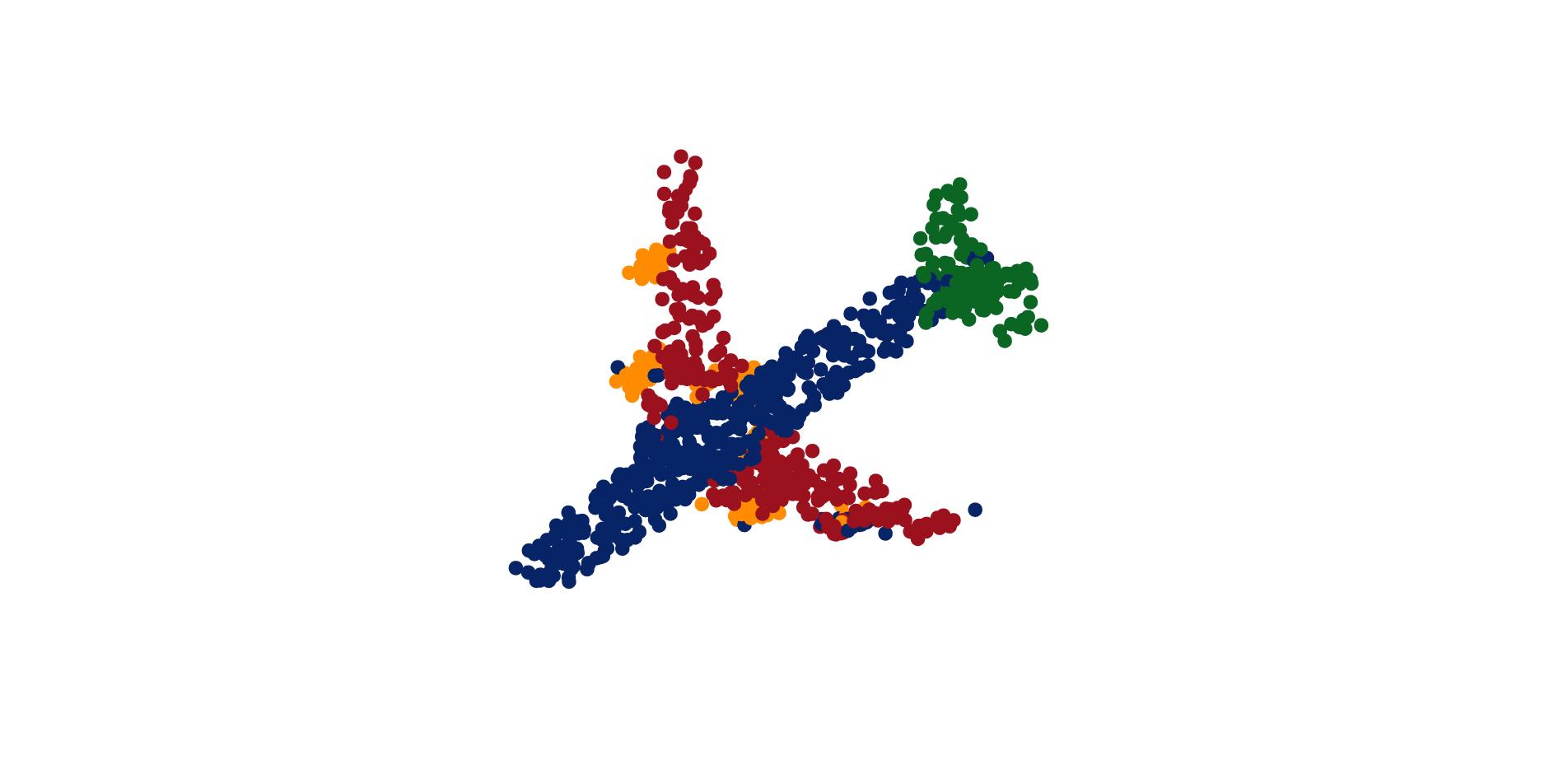}
    \includegraphics[trim={15cm 0 15cm 0},clip,width=0.136\linewidth]{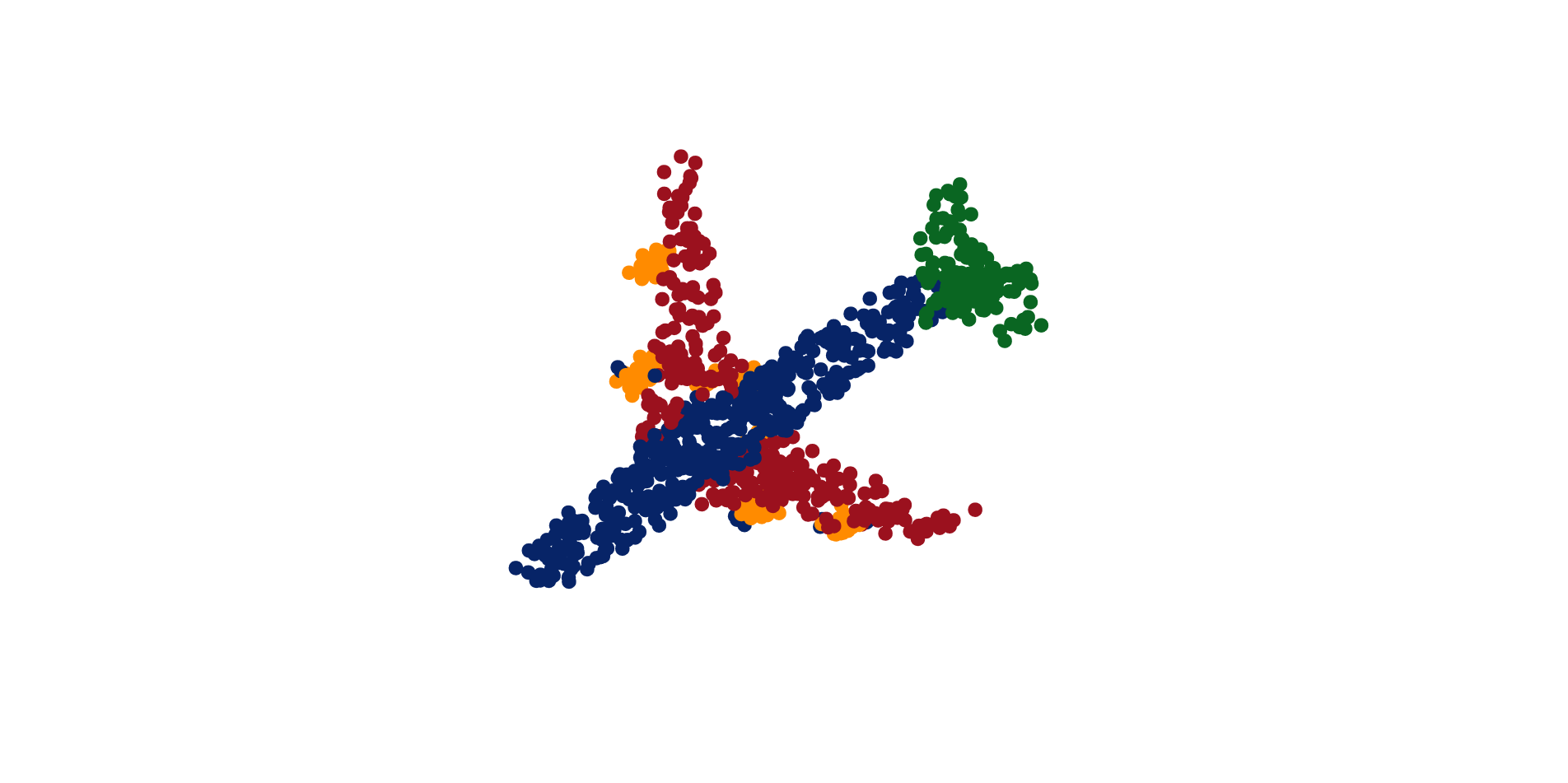}
    \includegraphics[trim={15cm 0 15cm 0},clip,width=0.136\linewidth]{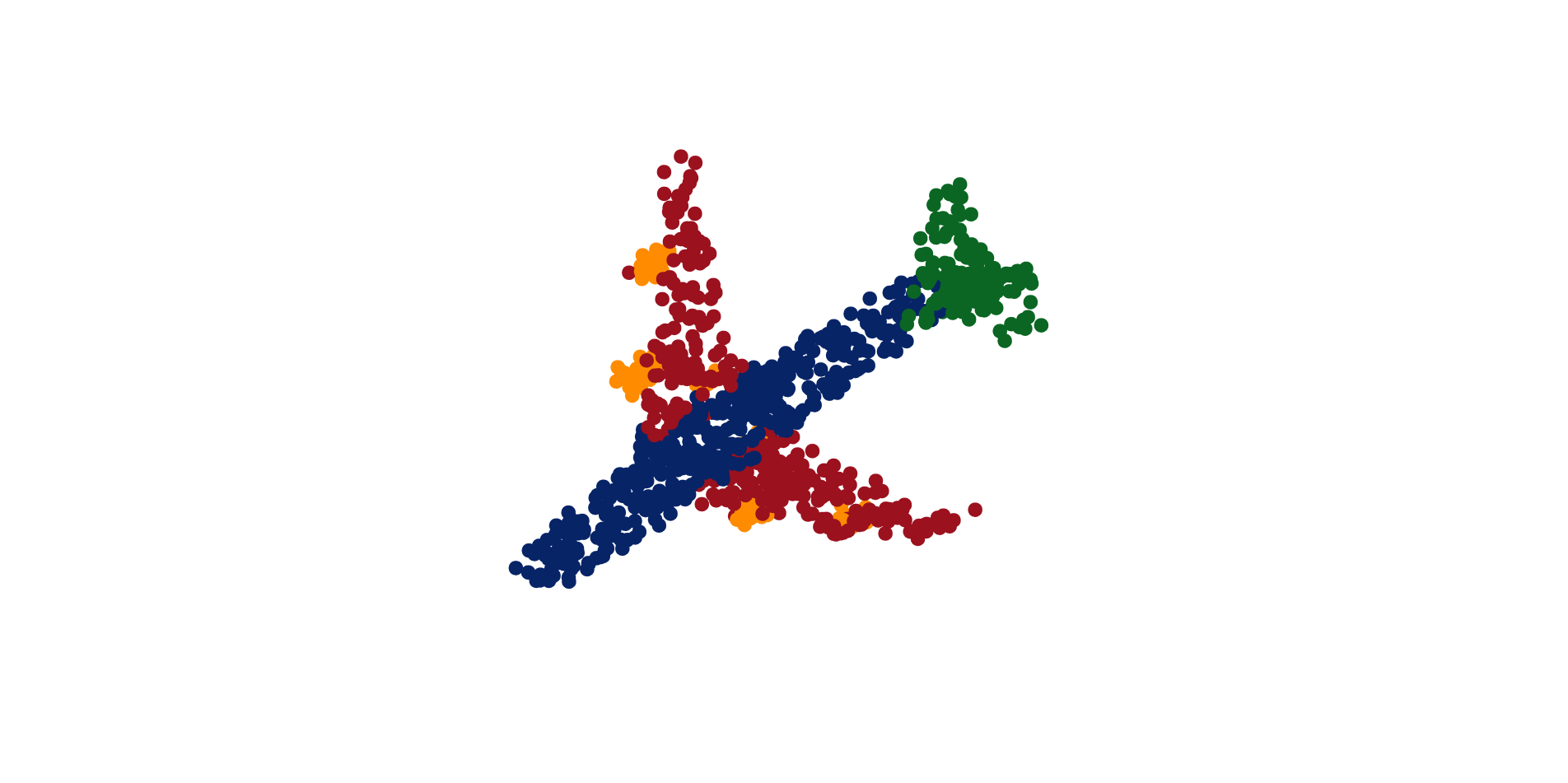}
    \includegraphics[trim={15cm 0 15cm 0},clip,width=0.136\linewidth]{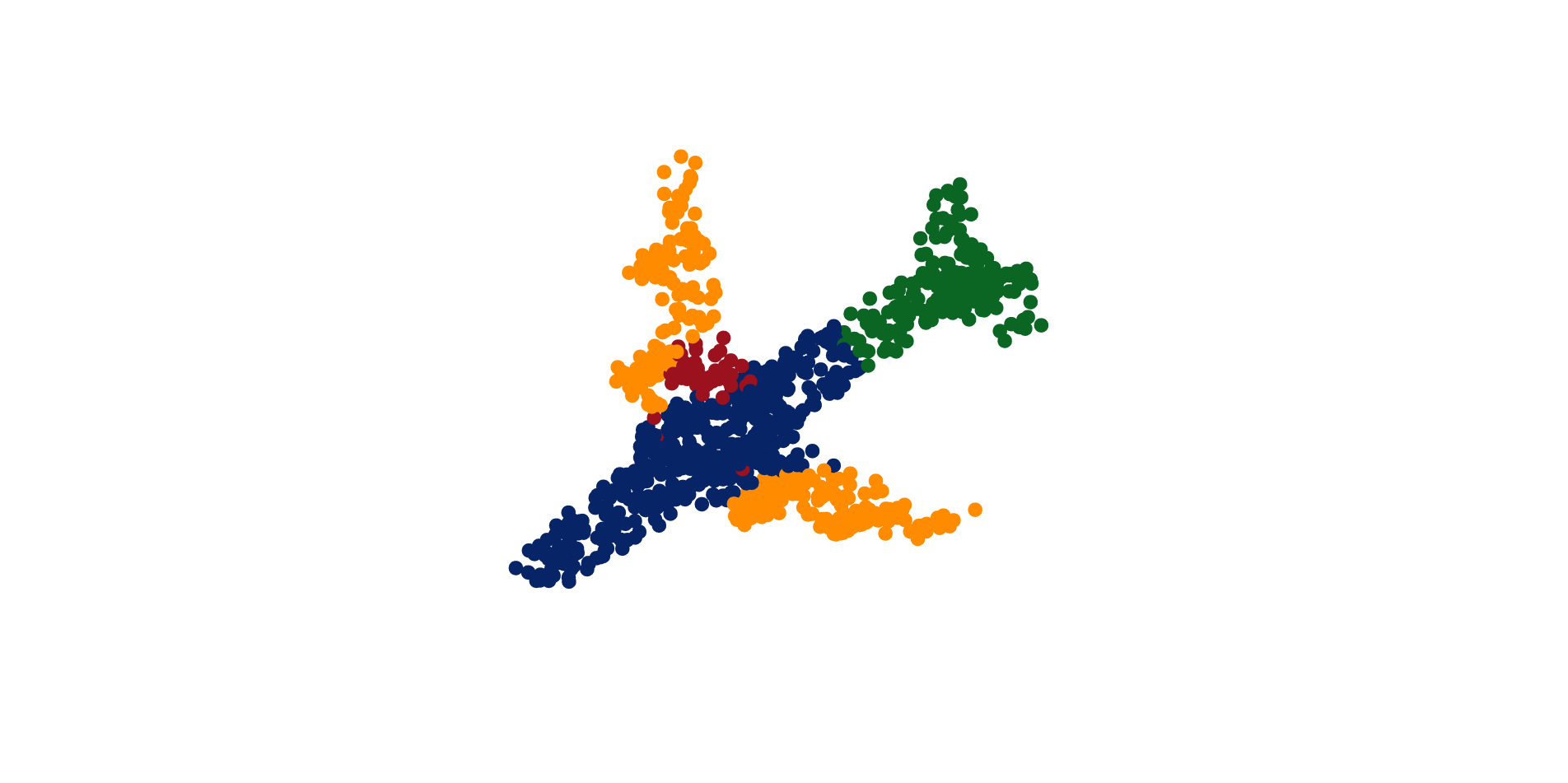}
    \includegraphics[trim={15cm 0 15cm 0},clip,width=0.136\linewidth]{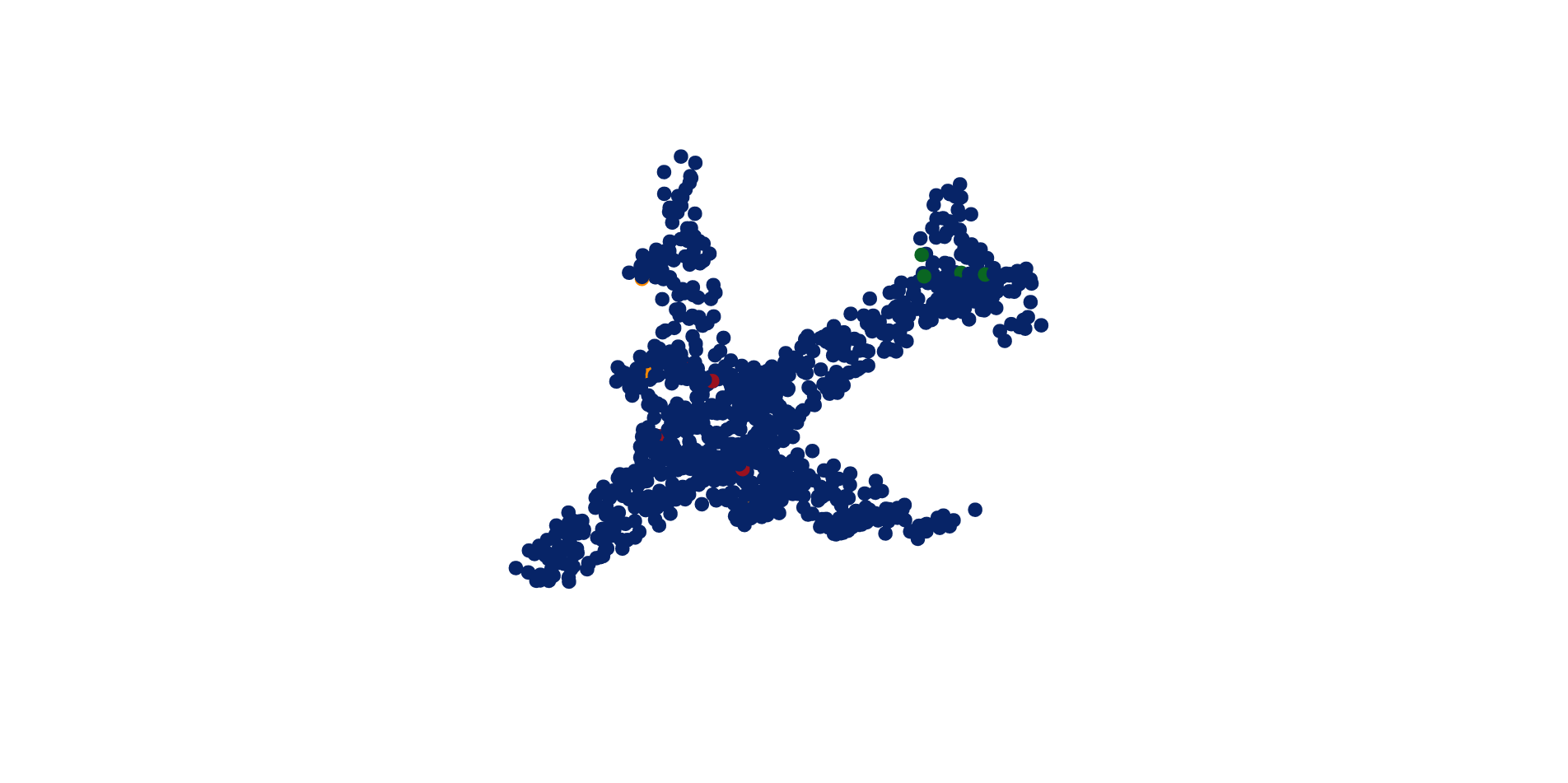}\\
    \includegraphics[trim={15cm 0 15cm 0},clip,width=0.136\linewidth]{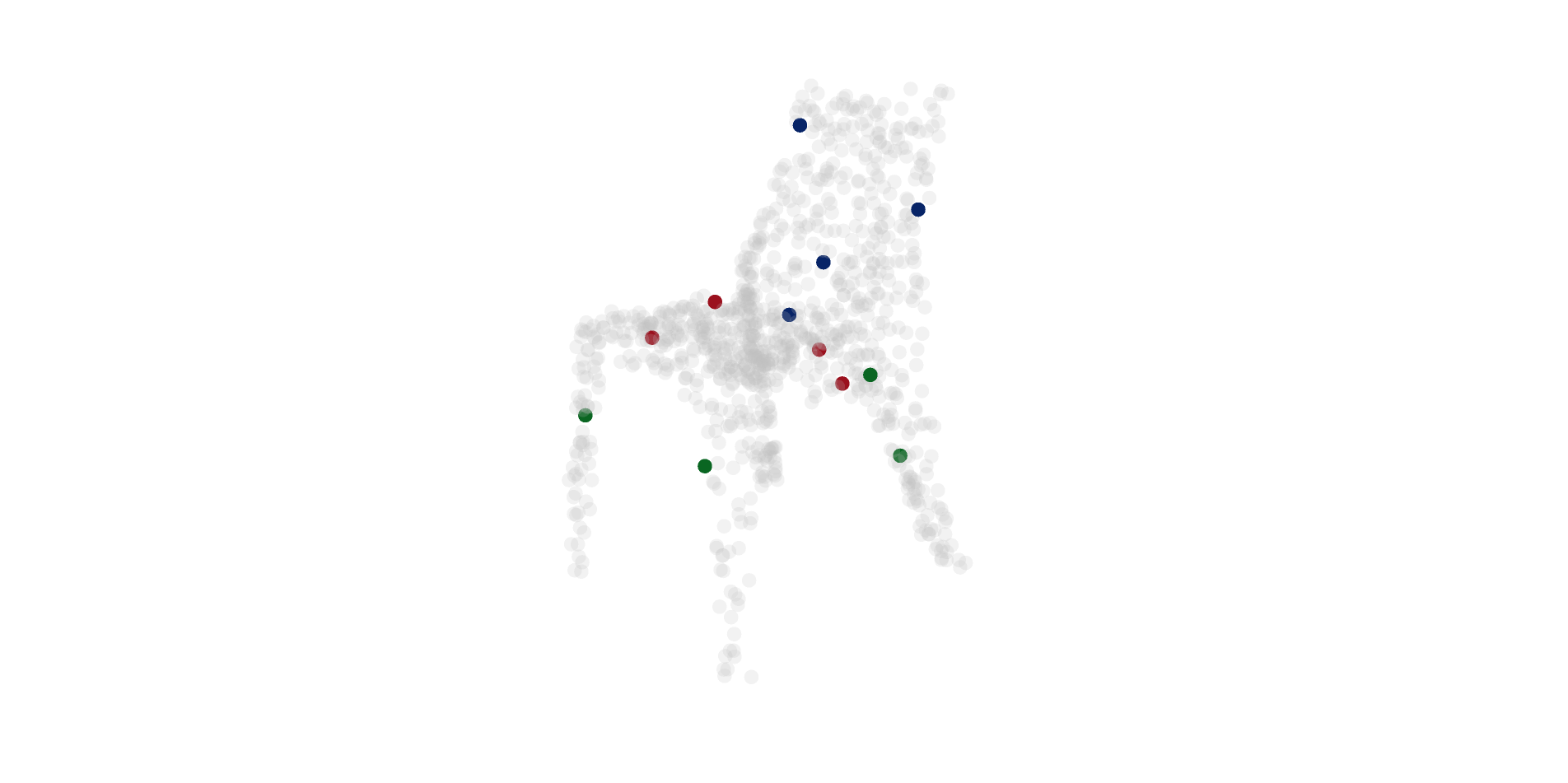}
    \includegraphics[trim={15cm 0 15cm 0},clip,width=0.136\linewidth]{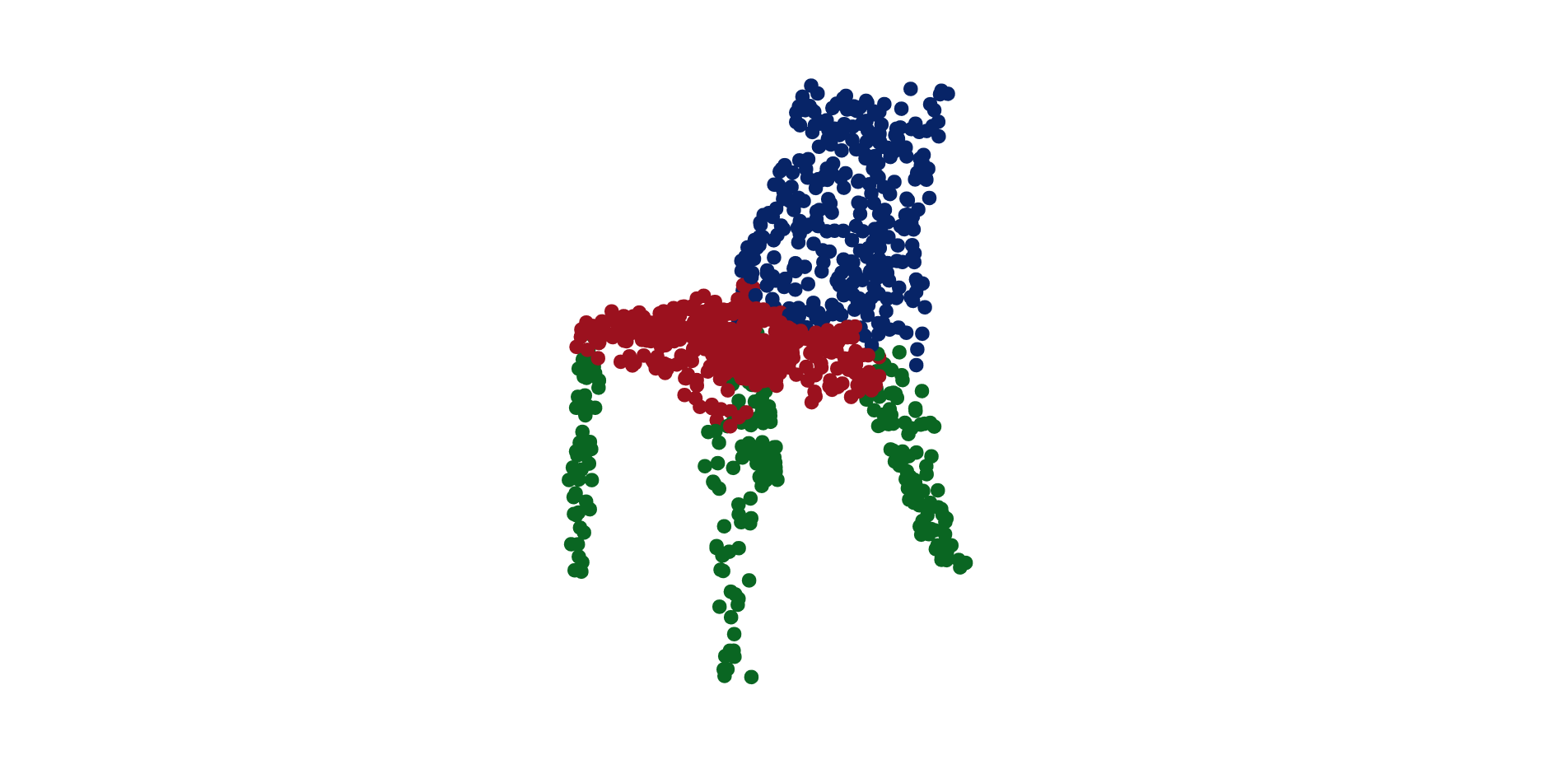}
    \includegraphics[trim={15cm 0 15cm 0},clip,width=0.136\linewidth]{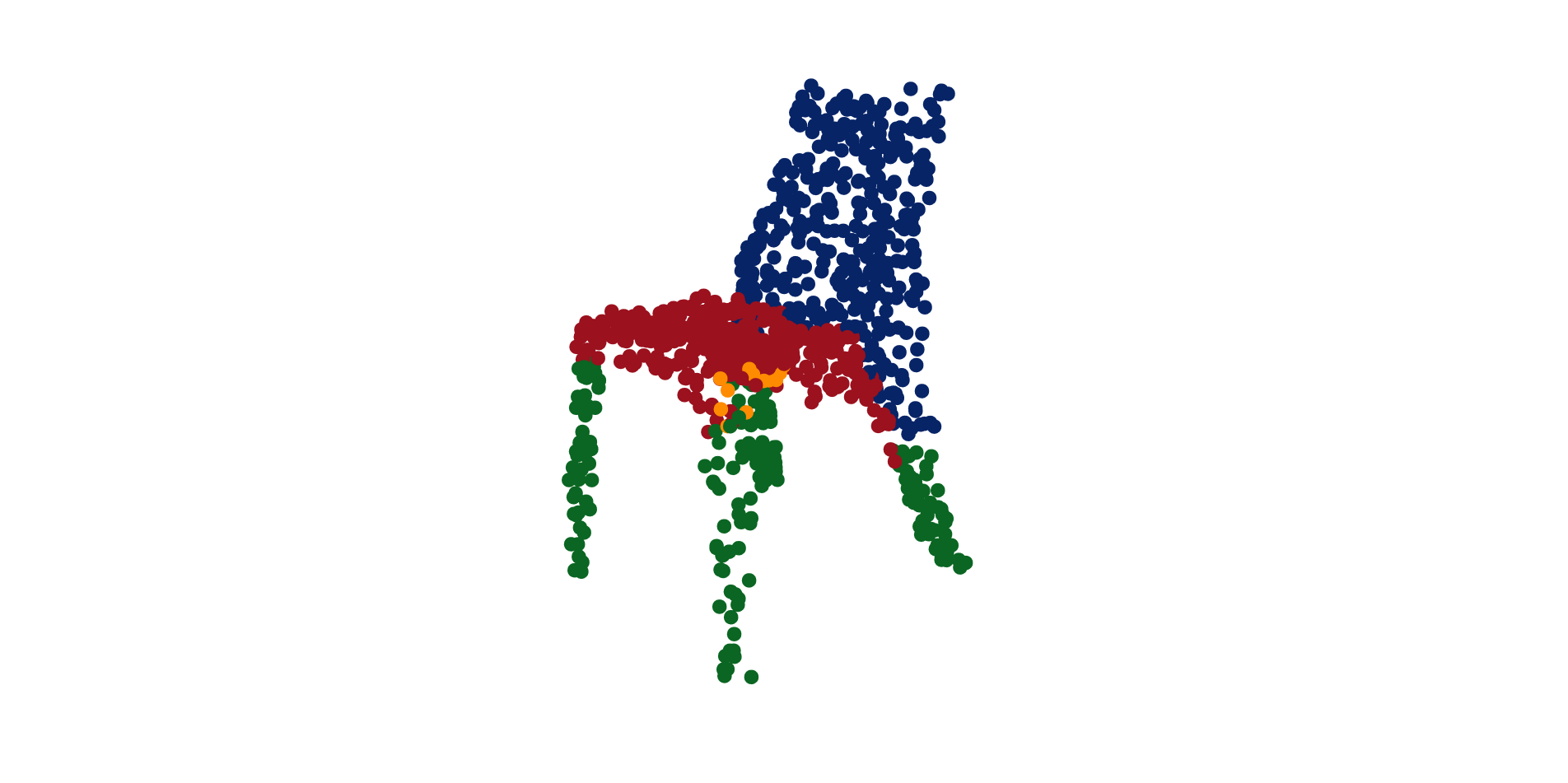}
    \includegraphics[trim={15cm 0 15cm 0},clip,width=0.136\linewidth]{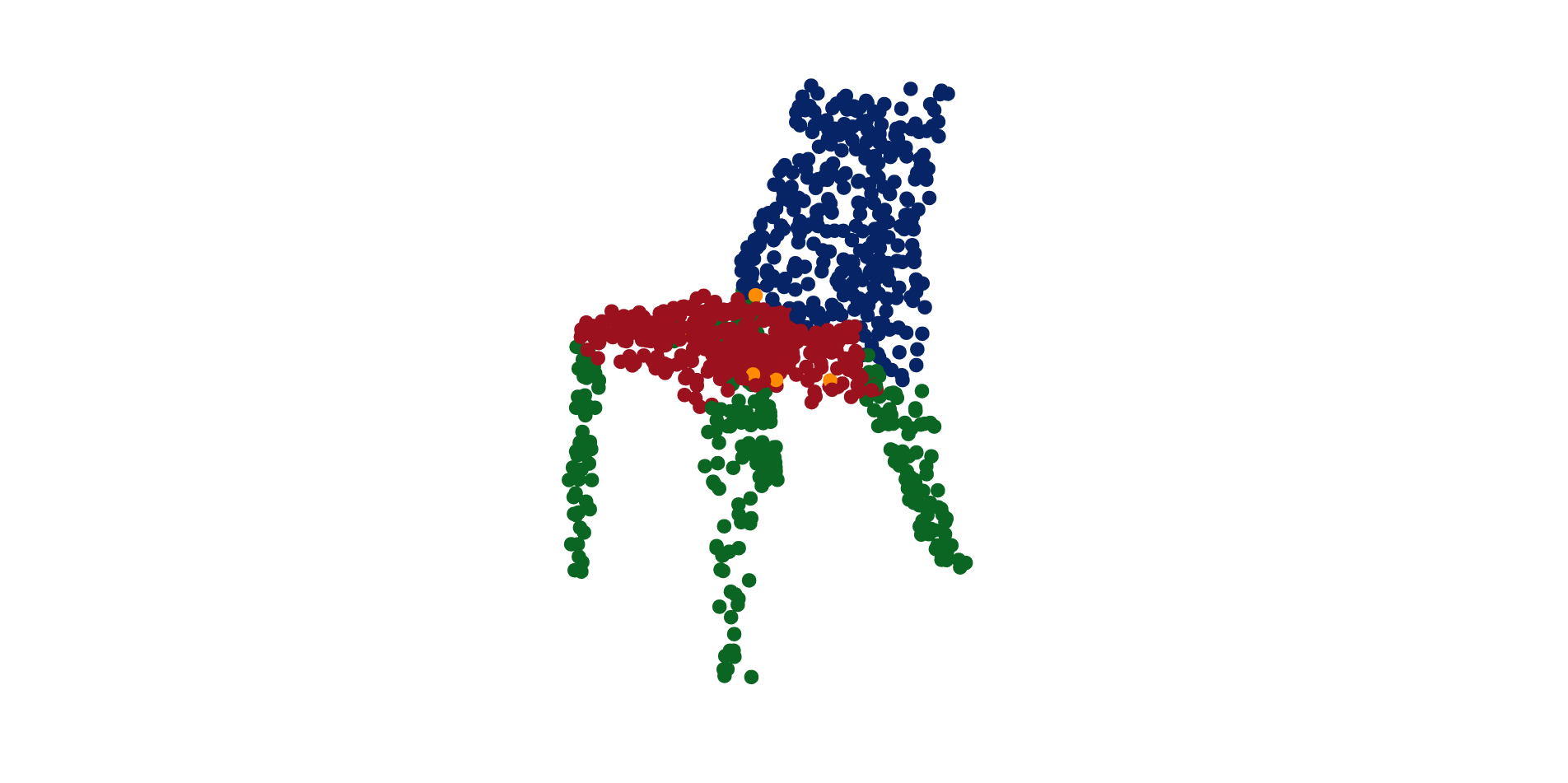}
    \includegraphics[trim={15cm 0 15cm 0},clip,width=0.136\linewidth]{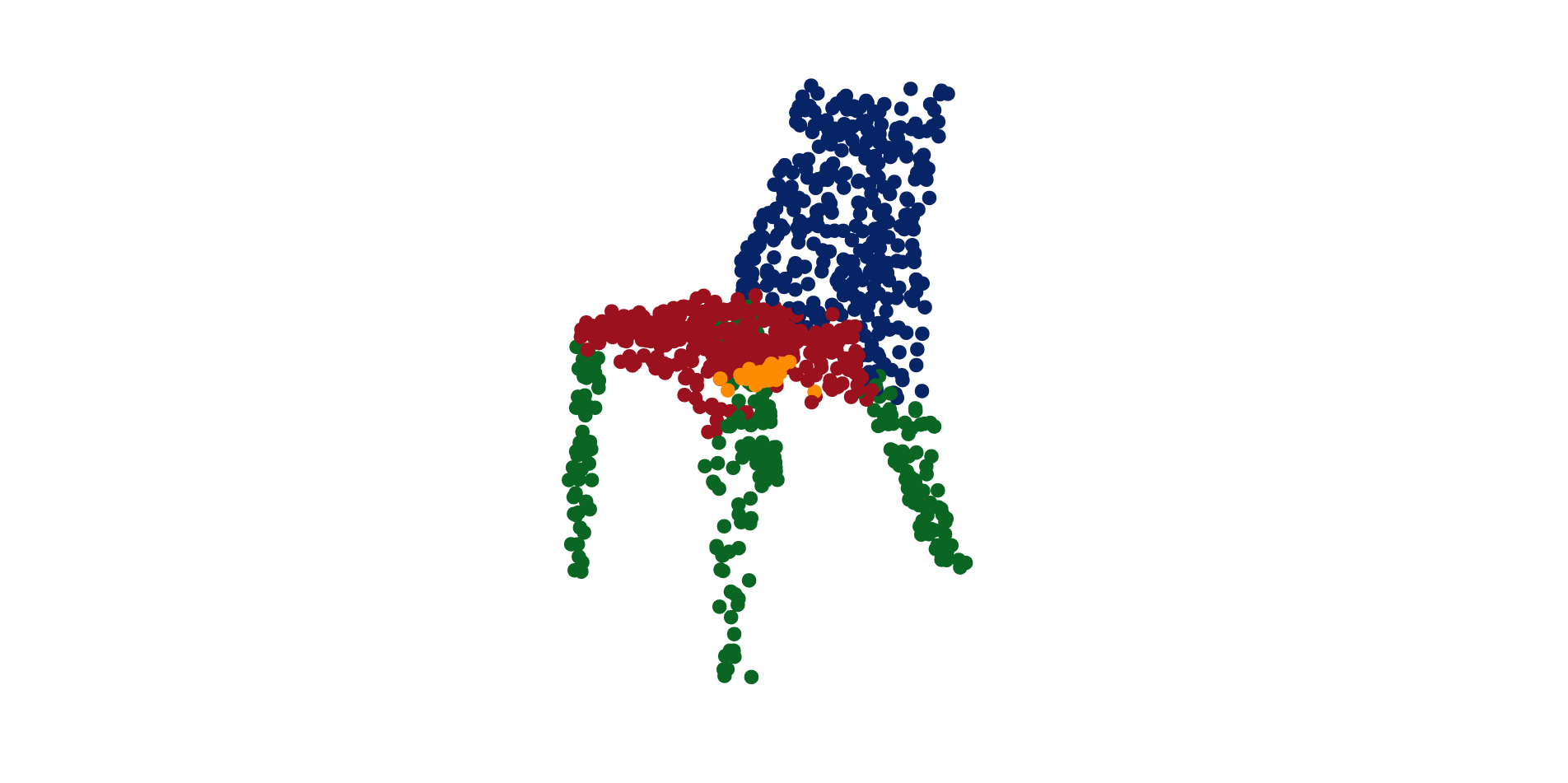}
    \includegraphics[trim={15cm 0 15cm 0},clip,width=0.136\linewidth]{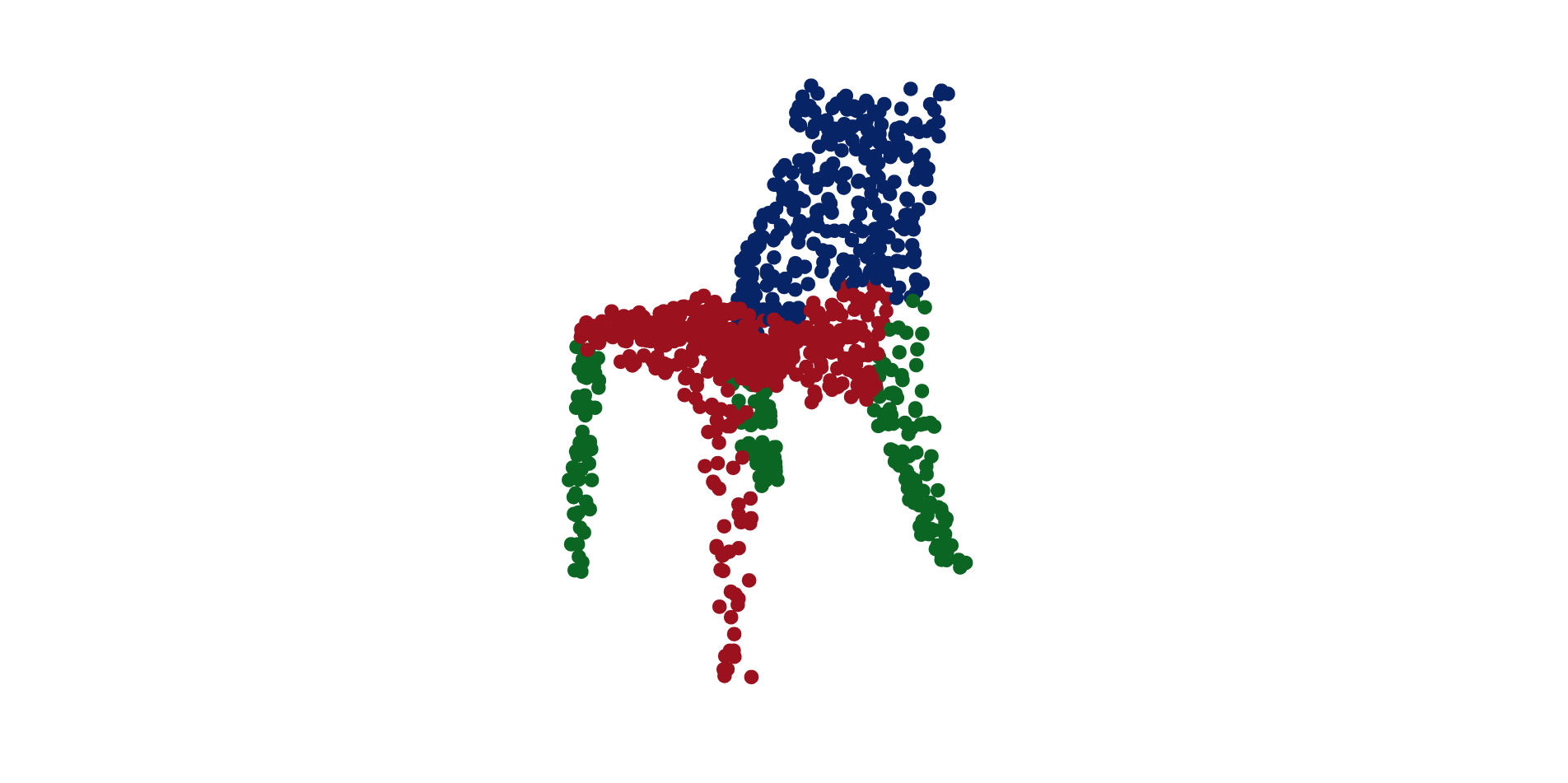}
    \includegraphics[trim={15cm 0 15cm 0},clip,width=0.136\linewidth]{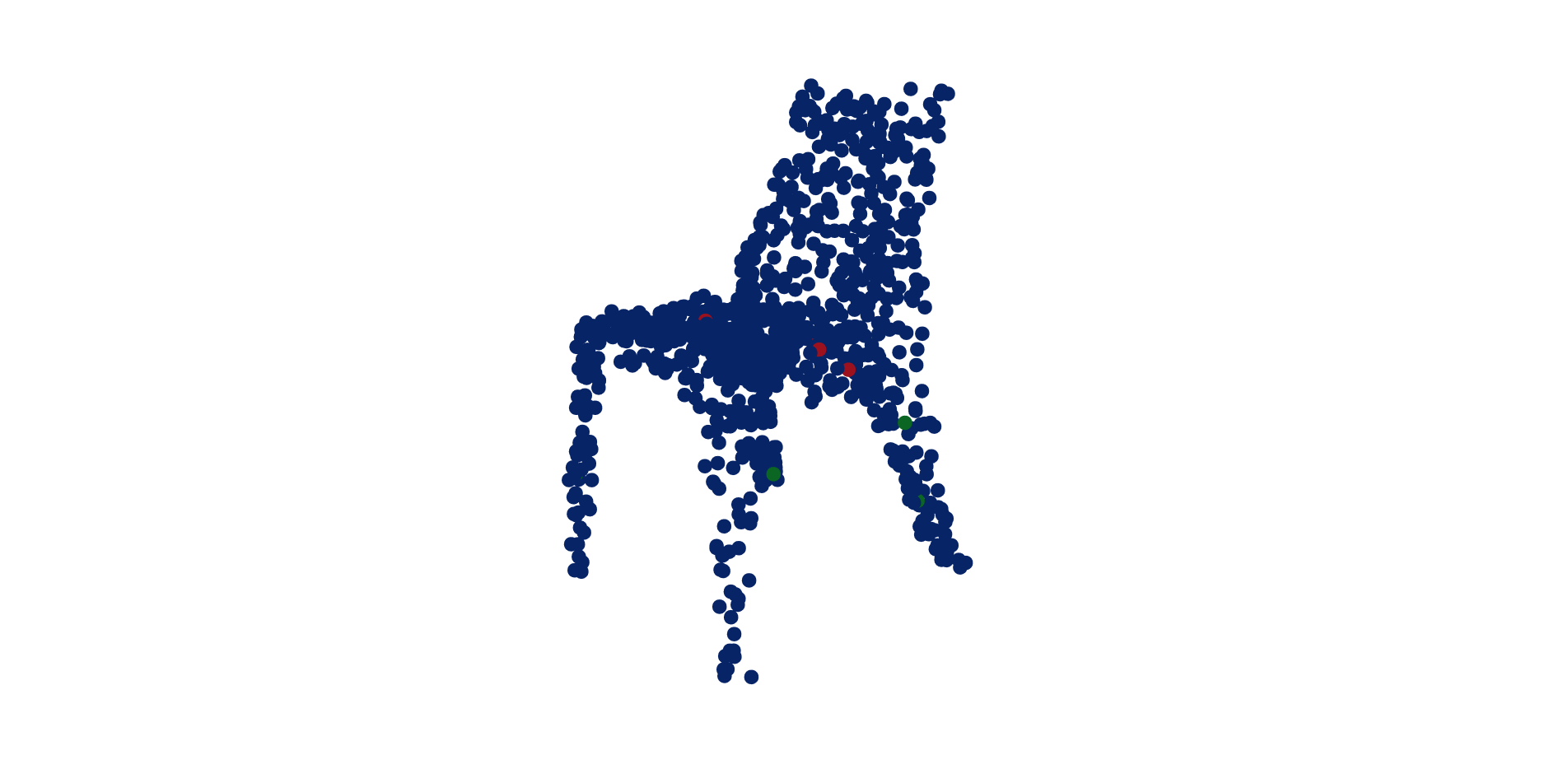}\\
    \includegraphics[trim={15cm 0 15cm 0},clip,width=0.136\linewidth]{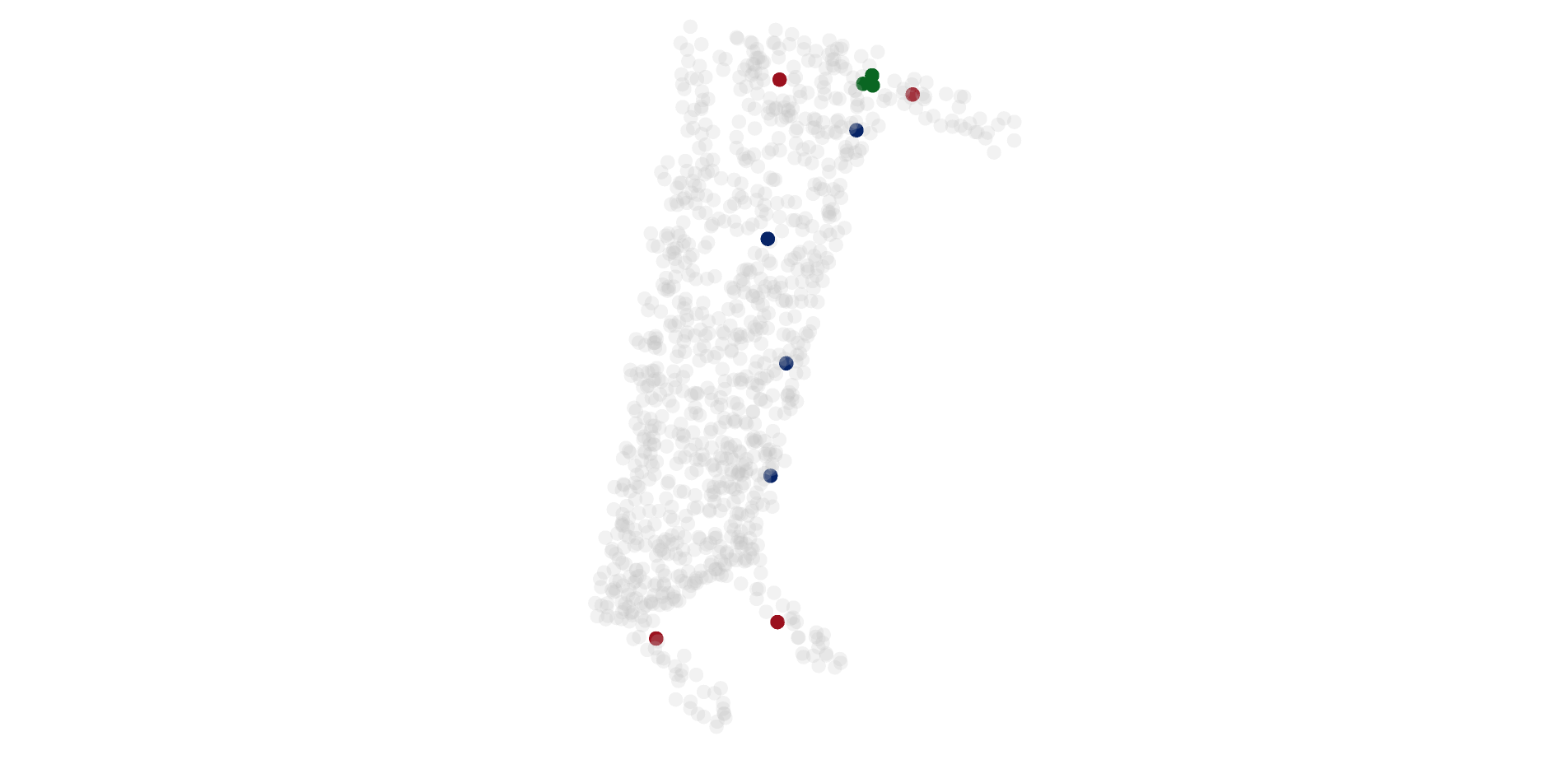}
    \includegraphics[trim={15cm 0 15cm 0},clip,width=0.136\linewidth]{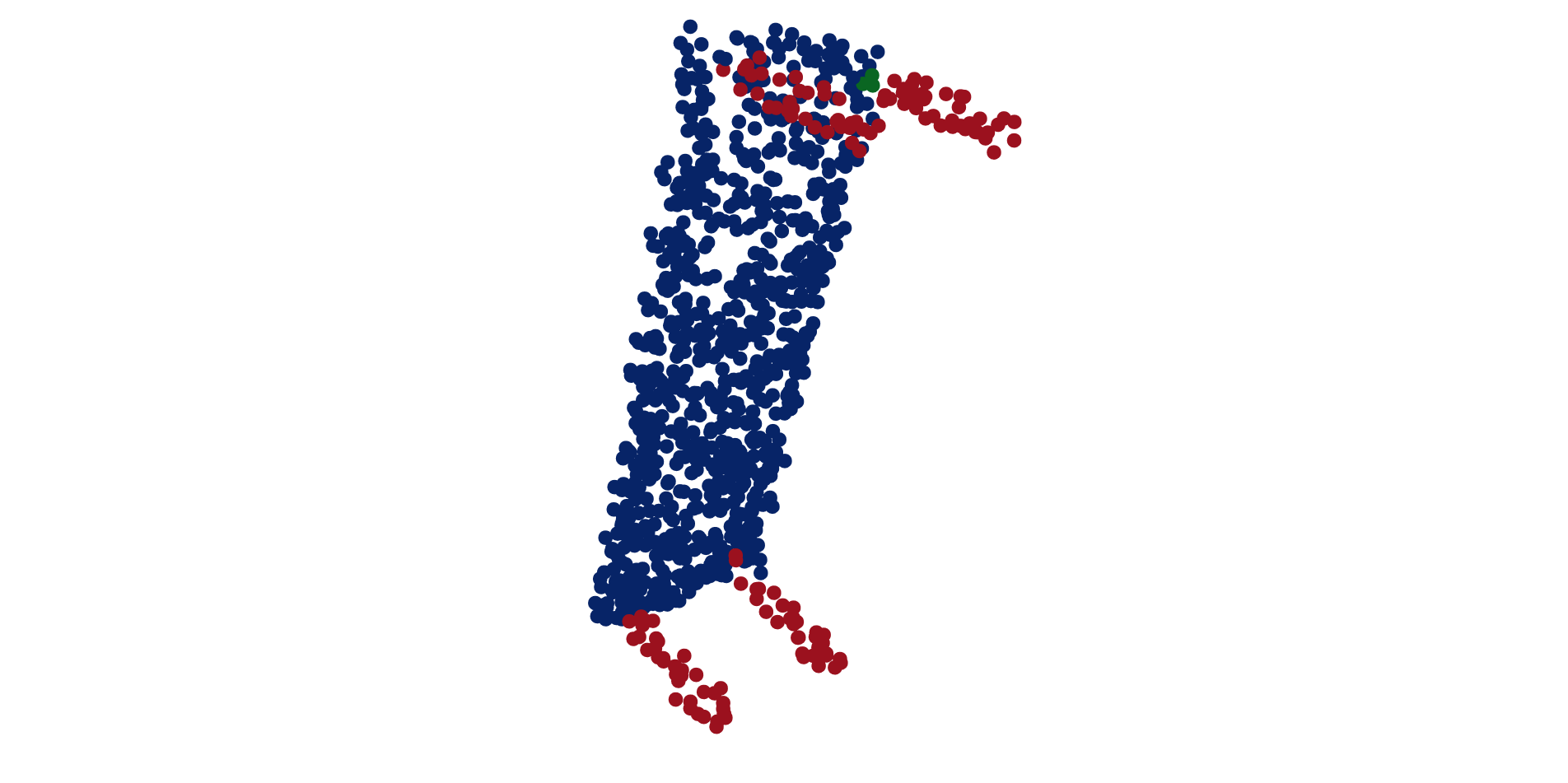}
    \includegraphics[trim={15cm 0 15cm 0},clip,width=0.136\linewidth]{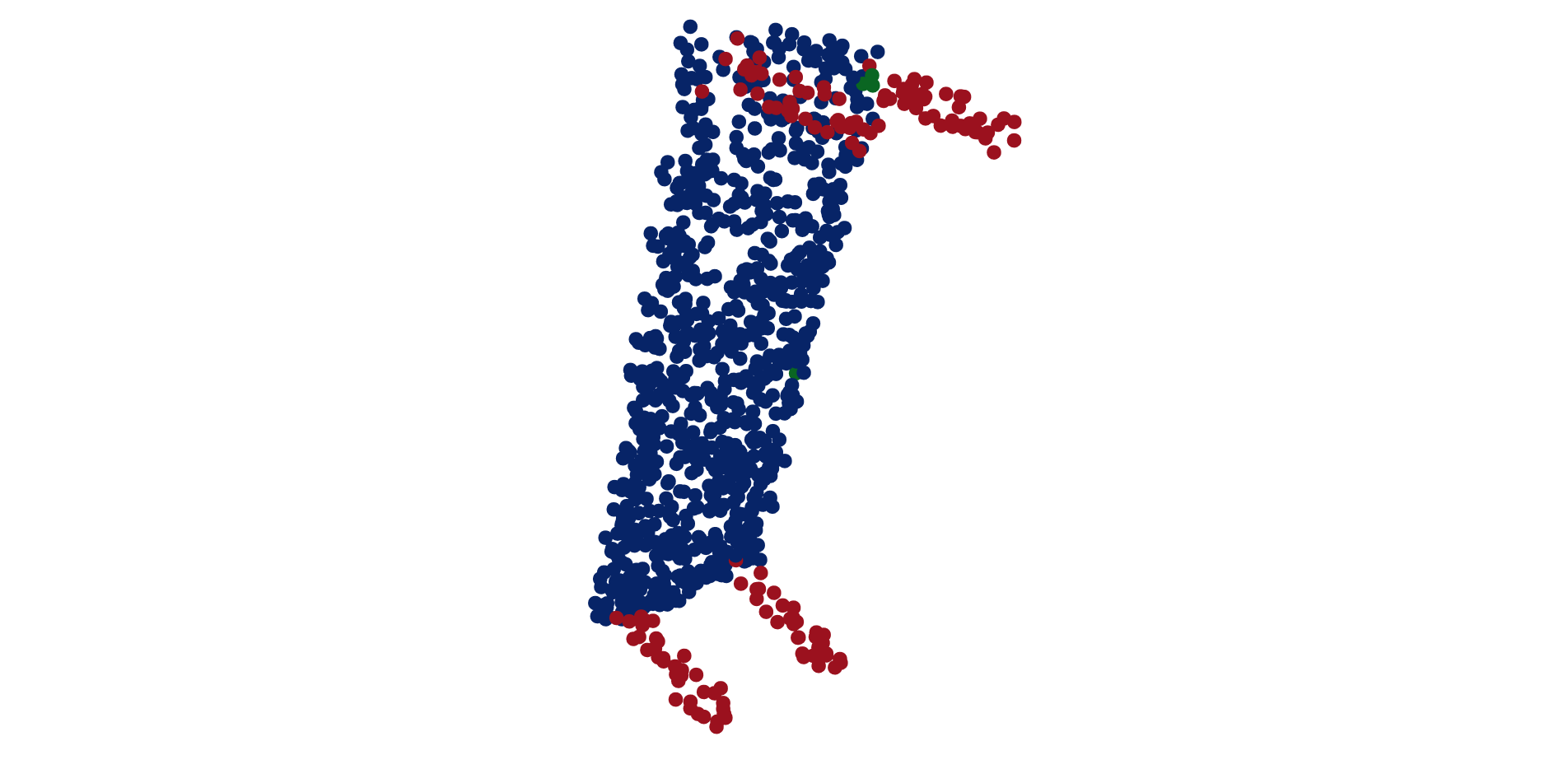}
    \includegraphics[trim={15cm 0 15cm 0},clip,width=0.136\linewidth]{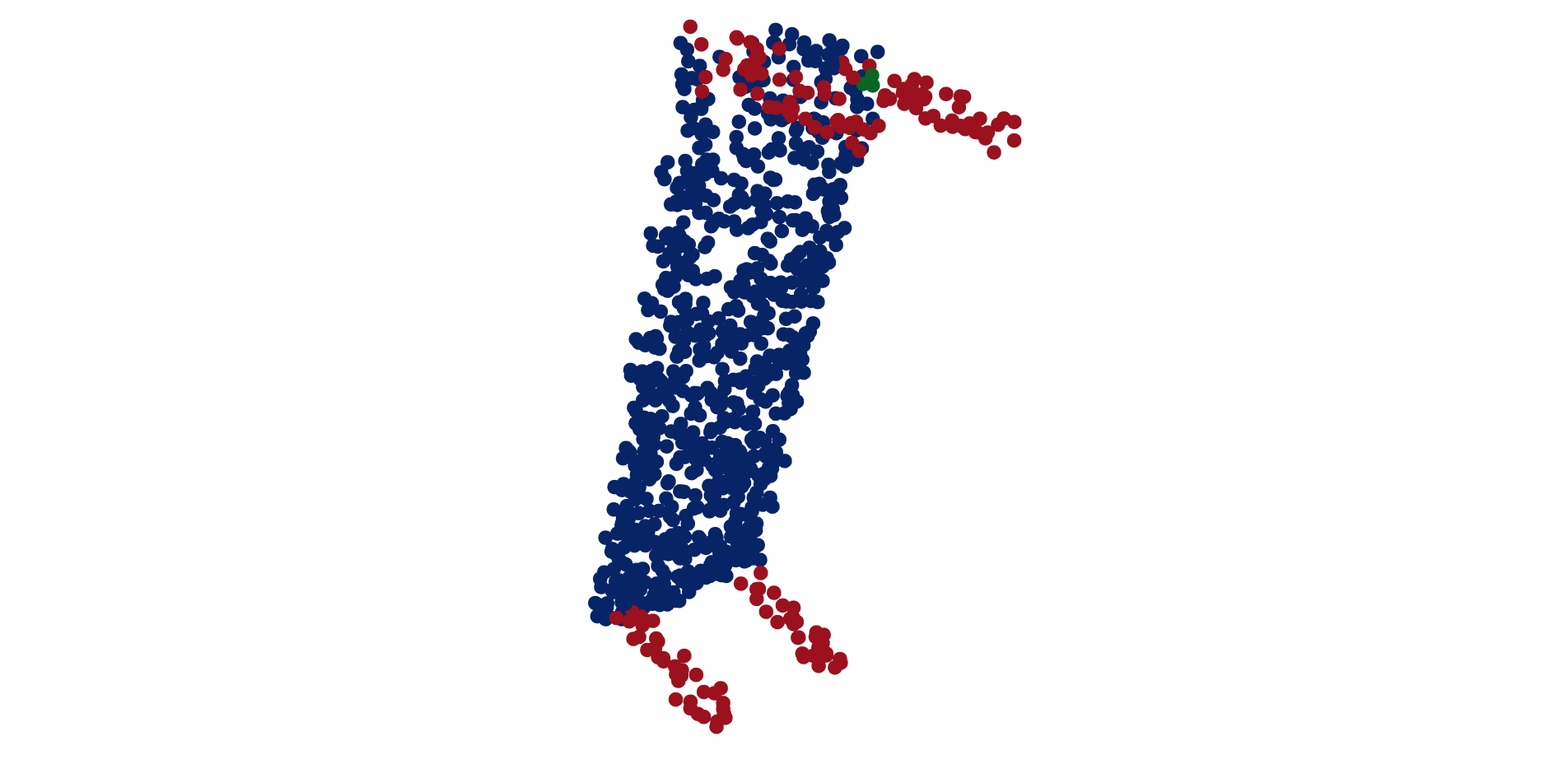}
    \includegraphics[trim={15cm 0 15cm 0},clip,width=0.136\linewidth]{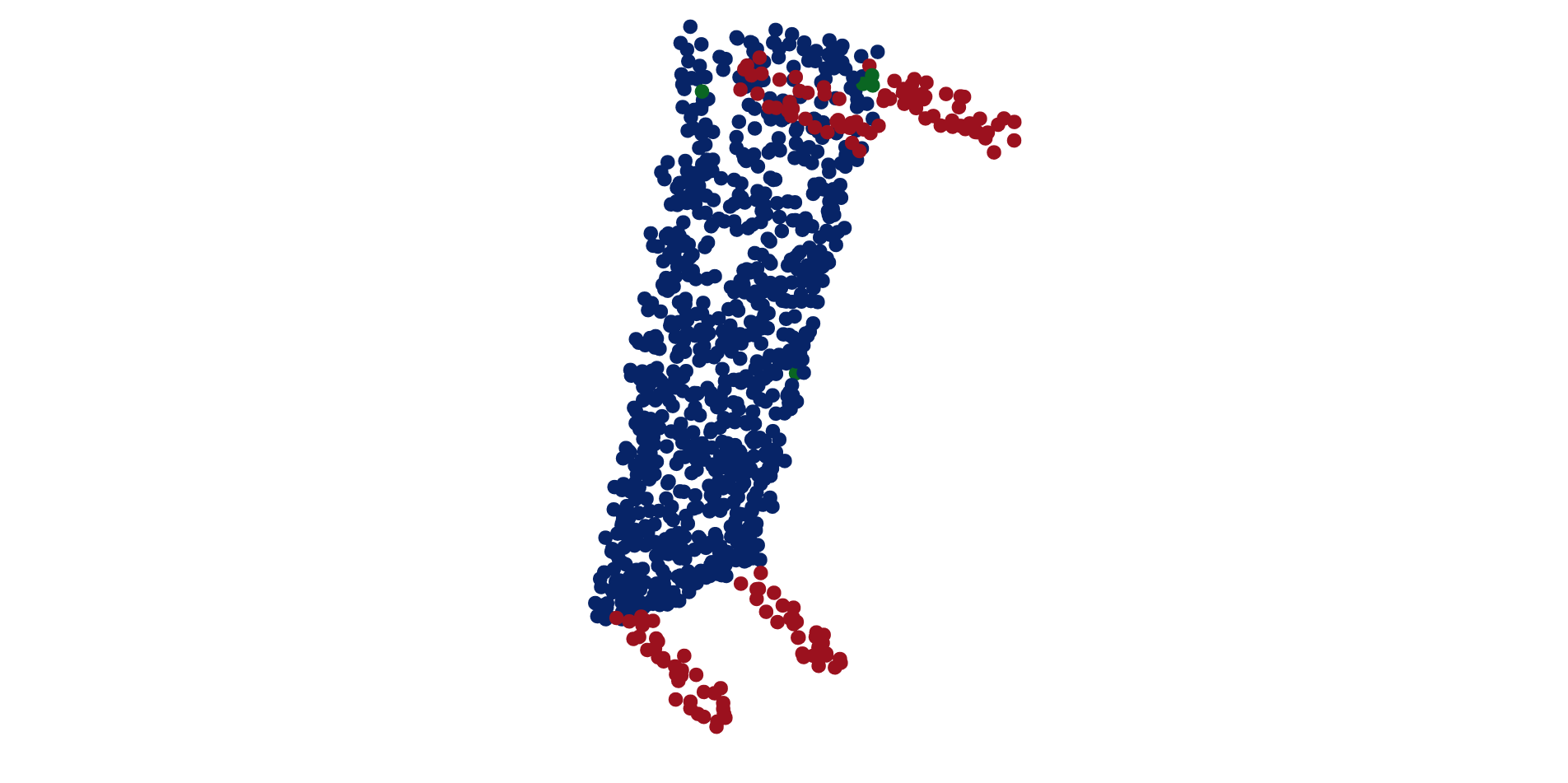}
    \includegraphics[trim={15cm 0 15cm 0},clip,width=0.136\linewidth]{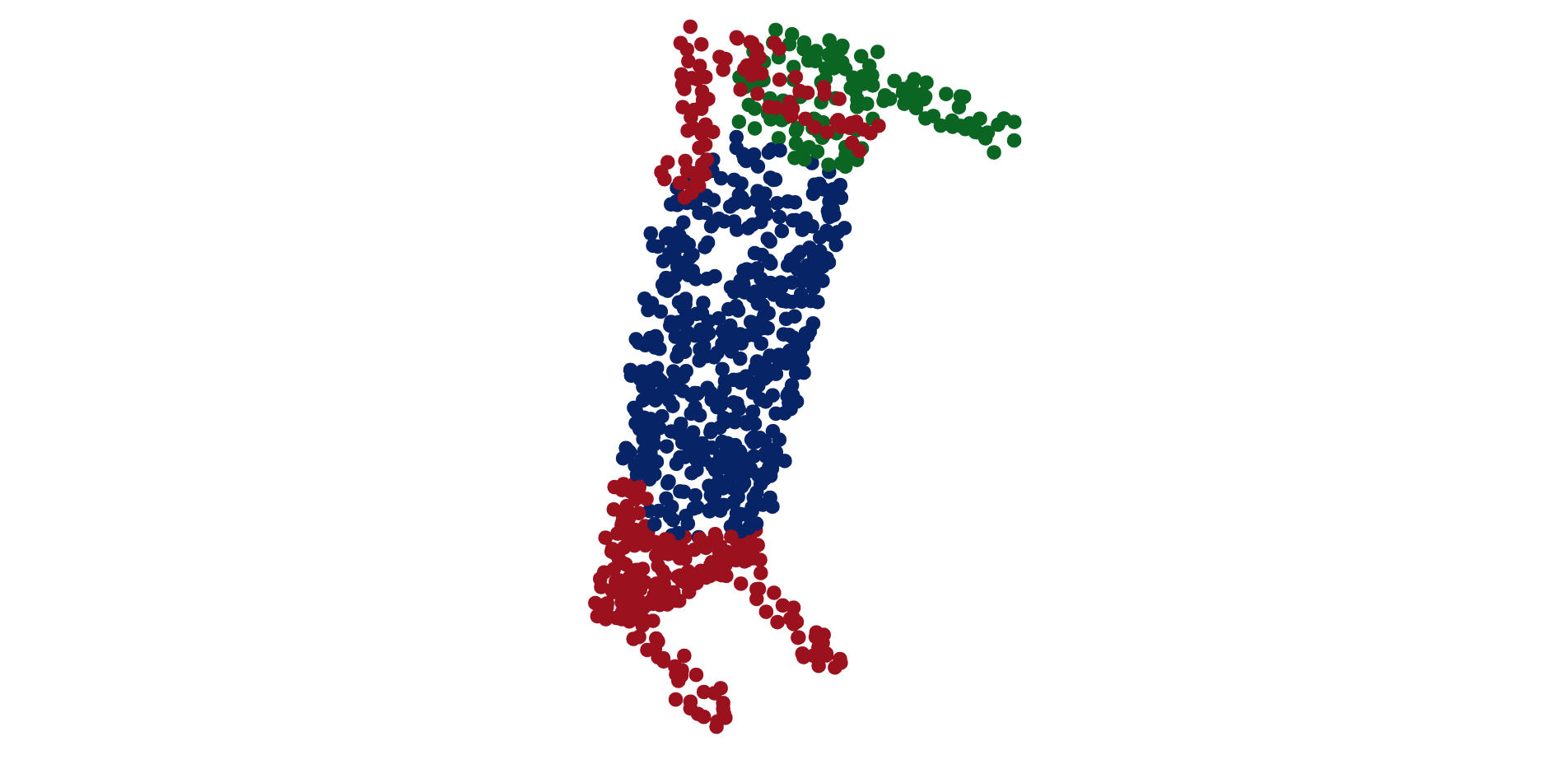}
    \includegraphics[trim={15cm 0 15cm 0},clip,width=0.136\linewidth]{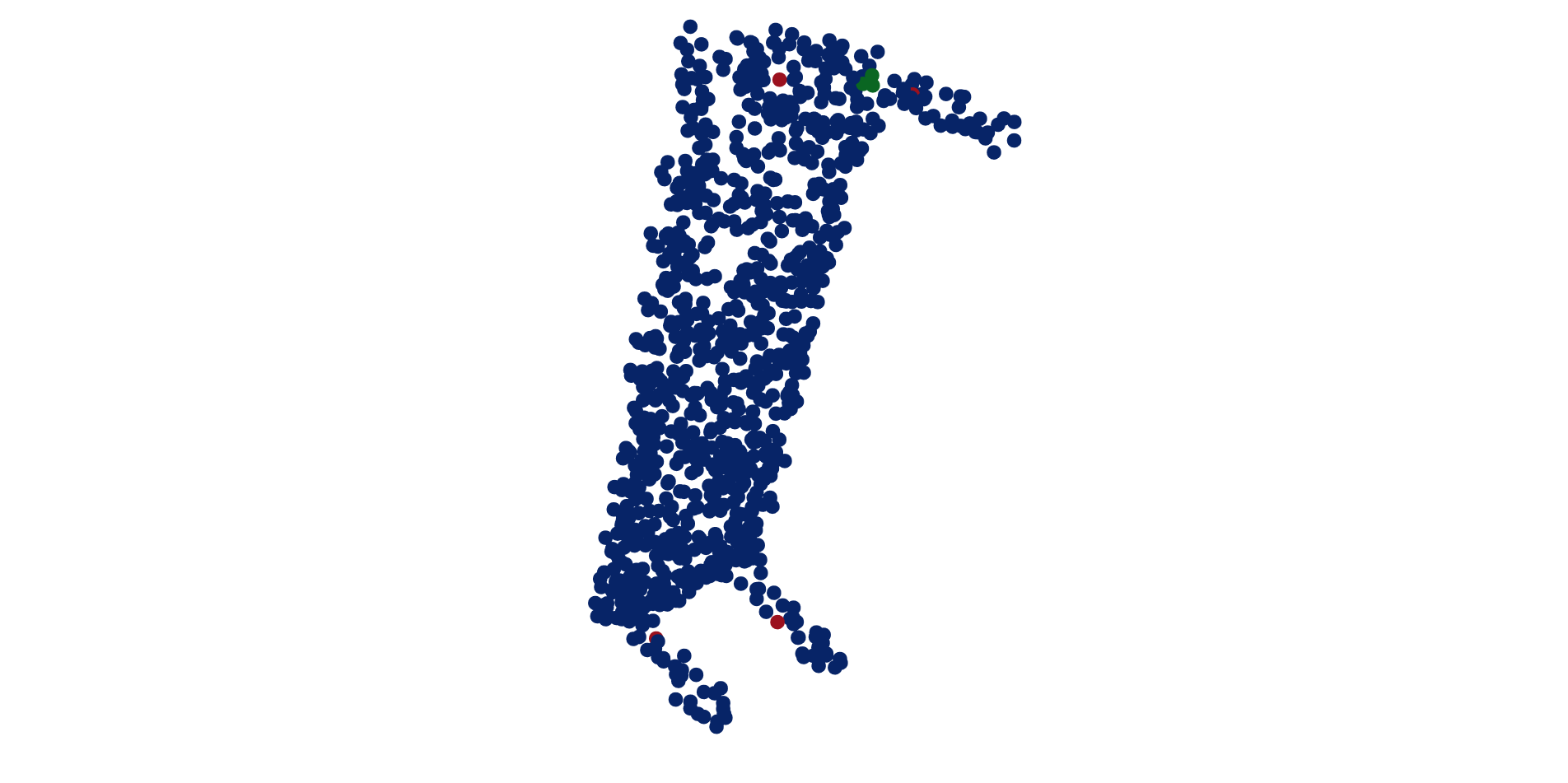}\\
    \parbox[b]{0.136\linewidth}{\centering Observed Data}
    \parbox[b]{0.136\linewidth}{\centering Ground-Truth}
    \parbox[b]{0.136\linewidth}{\centering Var-GNN}
    \parbox[b]{0.136\linewidth}{\centering ISS-GNN}
    \parbox[b]{0.136\linewidth}{\centering Prox-GNN}
    \parbox[b]{0.136\linewidth}{\centering LaplacianReg}
        \parbox[b]{0.136\linewidth}{\centering TikhonovReg}
    \caption{A qualitative comparison of different methods on the property completion problem on ShapeNet. Grey points in Observed Data indicate masked/unseen points. Neural approaches consistently outperform classical regularizers.}
    \label{fig:mask_task}
\end{figure*}

In the Property Completion problem, defined in Equation \eqref{eq:maskingOperator} and illustrated in Figure \ref{fig:gc}, for each graph, we randomly sample $nb$ nodes per class, and the selected nodes' labels are the observed data. For example, in Figure \ref{fig:mask_task}, we show an example where the property to be recovered are the node labels, and only $nb$ per class of them are available as input, besides possible meta-data such as coordinates. We present results with $nb=4$ in Table \ref{tab:mask_results}, and results on $nb=8, 16$ in Appendix \ref{appendix:additional_results}. We observe an accuracy-increasing trend is evident when increasing $nb$ and a consistently better performance offered by neural approaches.  
Also, we study the effect of meta-data, as shown in Table \ref{tab:meta_data_mask} (and in Appendix \ref{appendix:additional_results}), where a significant drop in performance is seen in the models when meta-data was not used.
 
\begin{table}[t]
\footnotesize
\centering
\caption{Property Completion performance (accuracy$\pm$standard deviation (\%)) on the ShapeNet and CLUSTER Datasets, with node budget per label per graph $nb=4$. The Cross-Entropy (CE) loss between the observed and reconstructed data is shown for reference.}
\label{tab:mask_results}
\begin{tabular}{llcccccc}
\toprule
Dataset & Model & Accuracy (\%)$\uparrow$  & Data Fit (CE$\downarrow$) \\ 
\midrule
\multirow{5}{*}{ShapeNet} 
& LaplacianReg & 74.90$\pm$0.00 & 3.87$\pm$0.00 \\ 
 & TikhonovReg & 6.91$\pm$0.00 & 3.71$\pm$0.00 \\ 
\cmidrule(r){2-4}
 & Var-GNN & 89.16$\pm$0.31 &  0.24$\pm$0.08  \\
 & ISS-GNN & 89.59$\pm$0.53 & 0.27$\pm$0.02 \\ 
 & Prox-GNN & 89.33$\pm$0.23 & 2.94$\pm$2.8$\cdot$10$^{-9}$ \\
\midrule
\multirow{6}{*}{CLUSTER} 
  & LaplacianReg & 70.13$\pm$0.00 & 1.78$\pm$0.00 \\
  & TikhonovReg & 34.64$\pm$0.00 & 1.12$\pm$0.00 \\
  \cmidrule(r){2-4}
   & Var-GNN & 88.89$\pm$0.51 & 1.04$\pm$4.9$\cdot 10^{-9}$ \\
  & ISS-GNN & 57.08$\pm$7.36 & 1.04$\pm$1.6$\cdot 10^{-8}$ \\ 
 & Prox-GNN & 53.58$\pm$2.58 & 1.04$\pm$0.00 \\
\bottomrule
\end{tabular}%
\end{table}

\begin{table}[t]
\footnotesize
\centering
\caption{The impact of meta-data within different networks, on the Property Completion task on the ShapeNet datasets with node budget per label per graph $nb=4$. The CE is between the observed and reconstructed data.}
\label{tab:meta_data_mask}
\begin{tabular}{llcc}
\toprule
Meta-data & Model & Accuracy (\%)$\uparrow$ & Data Fit (CE$\downarrow$)  \\
\midrule
Yes & \multirow{2}{*}{Var-GNN} & 89.16$\pm$0.31 & 0.24$\pm$0.08 \\
No &  & 26.89$\pm$0.02 & 0.37$\pm$0.02  \\
\hline
Yes & \multirow{2}{*}{ISS-GNN} &  89.59$\pm$0.53 & 0.27$\pm$0.02\\
No &  & 26.45$\pm$0.02 &  3.01$\pm 1.0\cdot 10^{-6}$ \\
\hline
Yes & \multirow{2}{*}{Prox-GNN} & 89.33$\pm$0.23 & 2.94$\pm2.8 \cdot 10^{-9}$  \\
No &  & 26.96$\pm$0.01 & 2.94$\pm4.1 \cdot 10^{-9}$  \\
\bottomrule
\end{tabular}%
\end{table}

\begin{table}[t]
\footnotesize
\centering
\caption{Inverse Source Estimation accuracy$\pm$standard deviation (\%) on CLUSTER dataset, with $k=16$ diffusion steps. CE is between the observed and reconstructed data.}
\label{tab:deblurring_results_cluster3}
\begin{tabular}{lcc}
\toprule
 Method & Accuracy (\%)$\uparrow$ & Data Fit (CE$\downarrow$) \\
\midrule
 LaplacianReg & 26.28$\pm$0.00& 1.75$\pm$0.00 \\
 TikhonovReg & 26.28$\pm$0.00 & 1.76$\pm$0.00 \\
\midrule
 Var-GNN &  82.36$\pm$0.20 & 1.75$\pm1.8 \cdot 10^{-9}$ \\
 ISS-GNN & 52.31$\pm$4.18 & 1.75$\pm3.1\cdot 10^{-9}$ \\
 Prox-GNN & 44.48$\pm$3.63 & 1.75$\pm3.8 \cdot 10^{-9}$ \\
\bottomrule
\end{tabular}%
\end{table}

 \subsection{Inverse Source Estimation}

We now discuss the results for the Inverse Source Estimation problem, which is described by the forward operator in Equation \eqref{eq:sourceProblem} and Figure \ref{fig:sl}. 
In this problem, node states {\bfx} are diffused by $k$ applications of the adjacency matrix where $k=4, 8, 16$, and their diffused result is the observed data. As $k$ grows, the problem becomes harder. The results with $k=16$ are reported in on the CLUSTER dataset in Table \ref{tab:deblurring_results_cluster3}, and on the CPOX dataset in Table \ref{tab:deblurring_results_cpox}, with additional results in Appendix \ref{appendix:additional_results}.  
We note that in the case of inverse source estimation, our ISS-GNN becomes similar to \citet{huang2023twostage}.
Our results indicate that all neural methods achieve better performance than classical methods.

\begin{table}[t]
\footnotesize
\centering
\caption{Inverse Source Estimation normalized-mean-squared error (nMSE) of the error and data-fit on the Chickenpox-Hungary dataset, with $k=16$ diffusion steps.  
}
\label{tab:deblurring_results_cpox}
\begin{tabular}{lcc}
\toprule
 Method  & nMSE$\downarrow$ & Data Fit (nMSE$\downarrow$) \\
\midrule
  LaplacianReg & 0.84$\pm$0.00 & 0.048$\pm$0.00 \\
 TikhonovReg  & 0.84$\pm$0.00 & 0.051$\pm$0.00 \\
 \midrule
   Var-GNN & 0.59$\pm$0.01 & $ 5.9 \cdot 10^{-11} \pm$0.00 \\
  ISS-GNN  & 0.59$\pm 8.4\cdot 10^{-5}$ & 5.5$\cdot 10^{-11}\pm$0.00 \\
  Prox-GNN & 0.73$\pm$0.00 & $1.0   \cdot 10^{-5}\pm0.00$ \\
\bottomrule
\end{tabular}%
\end{table}

\subsection{Inverse Graph Transport}
This Inverse Graph Transport is described by Equation \eqref{eq:graphTrsnportProblem} and by Figure \ref{fig:gtomo}. 
We study the performance of the different models for path lengths $pl=8, 16, 32$, meaning that the observed data at each node is an average of $pl$ features of nodes that lie on a path that starts from the respective node (as further explained in Appendix \ref{appendix:datasets}). 
The results with $pl=32$ are shown in Table \ref{tab:pathlen_results2}, indicating that neural models significantly outperform classical approaches, showing their problem efficacy at solving ill-posed problems. 

\begin{table}[t]
\footnotesize
\centering
\caption{Inverse Graph Transport normalized-mean-squared-error (nMSE) on the METR-LA dataset with $pl=32$.}
\label{tab:pathlen_results2}
\begin{tabular}{lcccccc}
\toprule
 Model &nMSE$\downarrow$ & Data Fit (nMSE$\downarrow$) \\
\midrule
   LaplacianReg  & 0.61$\pm$0.00 &  0.11$\pm$0.00 \\
 TikhonovReg & 0.36$\pm$0.00 & 0.04$\pm$0.00 \\
 \midrule
   Var-GNN  & 0.004 $\pm2.0 \cdot 10^{-5}$ & $8.2 \cdot 10^{-6}\pm$0.00\\
  ISS-GNN & 0.21$\pm5.9 \cdot 10^{-5}$ & 0.004$\pm7.9 \cdot 10^{-6}$ \\
  Prox-GNN & 0.21$\pm3.5 \cdot 10^{-3}$ &  0.003$\pm1.4 \cdot 10^{-4}$ \\
\bottomrule
\end{tabular}%
\end{table}

\subsection{Edge Property Recovery}
We study a \emph{nonlinear} inverse problem, differently than the previously presented linear problems. Here, the goal is to recover edge weights given data that was diffused using these weights, as described in Equation \eqref{den} and by Figure \ref{fig:er}. Results on the superpixels CIFAR10 dataset are shown in Table \ref{tab:edge_rec_results}, where we see that classical methods struggle to offer a low relative error, compared with $~5$ times lower (better) error given by Var-GNN and ISS-GNN. The results on different GRIPs and datasets show the generality of the framework.

\begin{table}[t]
\footnotesize
\centering
\caption{Edge Property Recovery results on the CIFAR10 (superpixels) dataset.}
\label{tab:edge_rec_results}
\begin{tabular}{lccc}
\toprule
Model &  nMSE$\downarrow$ & Data Fit (nMSE$\downarrow$)  &\\
\midrule
  LaplacianReg & 0.598$\pm$0.00  &    0.0697$\pm$0.00\\ 
  TikhonovReg &0.626$\pm$0.00    &   0.0581$\pm$0.00\\ 
  \midrule
     Var-GNN & 0.114$\pm$0.03 &     0.073$\pm 2.4\cdot 10^{-5}$ \\
 ISS-GNN  & 0.122$\pm$0.02 & 0.078$\pm 3.9\cdot 10^{-5}$   \\
   Prox-GNN &   0.434$\pm$0.02 &   0.023$\pm 6.8\cdot 10^{-6}$ \\
\bottomrule
\end{tabular}
\end{table}

\section{Conclusions and Discussion}
\label{sec:conclusion}

In this paper, we propose a framework for Graph Inverse Problems. We show that while GRIPs can have very different forward problems, the recovery process can be addressed in a very similar fashion. We have experimented with a number of architectures for the solution of the problem, ranging from classical ones to learned ones. We observe that learning the regularization typically improves over unlearned regularization techniques. One important feature of GRIPs is the varying graph structure, and the second is the presence of meta-data. While meta-data is common for GRIPs, it is much less common for other inverse problems. We also show that using learned graph-based regularizations provides a natural way to include meta-data in the inversion process, leading to improved solutions. 

\newpage

\section*{Acknowledgments}
ME is funded by the Blavatnik-Cambridge fellowship, the Accelerate Programme for Scientific Discovery, and the Maths4DL EPSRC Programme.
\bibliography{biblio}


\clearpage

\appendix

\section{Related Works}
\label{sec:related_work}
The solution of inverse problems on graphs links between two fields, namely, Inverse Problems and Graphs. In this section, we provide an overview of the utilization of neural networks to solve inverse problems, followed by inverse problems on graphs.

\subsection{Solving Inverse Problems with Neural Networks}
\label{sec:related_IPnn}
Inverse problems are a crucial area in many scientific and engineering domains, where one aims to infer the underlying causes from observed effects. Traditional methods for solving inverse problems include variational approaches and regularization techniques, such as Tikhonov regularization,  Bayesian inference \cite{Tikhonov1977, Kaipio2006} and inverse scale-space techniques \cite{burger2006}. These methods often rely on explicit mathematical formulations and assumptions about the underlying models.
In recent years, neural networks have emerged as a powerful tool for addressing inverse problems, owing to their ability to learn complex mappings from data. Starting from the work \cite{adler2017solving}, which used proximal gradient descent, to related work  \cite{mardani2018neural,aggarwal2018modl,bai2020deep,ongie2020deep, Jin2017}, which used similar algorithms, deep networks have been used as priors and leveraged the powerful feature extraction capabilities to enhance reconstruction accuracy.
Recent work used a variational framework \cite{eliasof2023drip} to achieve similar goals.
More recently, diffusion models have been used to construct a prior for inverse problems \cite{chung2022diffusion}. While at inference, they perform very similar procedures as in the methods proposed above, they are trained very differently and not in an end-to-end fashion. We refer the reader for a comprehensive review of neural approaches for solving inverse problems to \citet{arridge2019solving}

\subsection{Inverse Problems on Graphs}
\label{sec:related_IPgraphs}
Compared to the solution of inverse problems on structured grids (e.g., 2D deblurring or 3D tomographic reconstruction), there is a much smaller body of work for the solution of inverse problems on graphs, which are unstructured in nature.
Traditional methods for graph inverse problems, do not include learnable networks and typically rely on optimization techniques and probabilistic models.
Similar to standard inverse problems, one of the foundational approaches in this domain is the use of graph regularization techniques, which incorporate prior knowledge about the graph structure to solve inverse problems more effectively \cite{Zhou2004, zhou2005regularization}. These methods have been widely used in semi-supervised learning and signal processing on graphs.
In recent years, the application of GNNs to inverse problems has been tailored to a particular application. Two recent cases are  node classification (see e.g. \cite{velickovic2018graph} and reference within)
and source estimation \cite{NEURIPS2023_46ab9d96, huang2023twostage}.
In recent work, the problem of source estimation
has been solved by using diffusion models as priors \cite{ling2024source}. This idea is similar to the use of diffusion models as priors in the context of inverse problems as discussed above.

\section{Broader Impact and Ethical Statement}
\label{appendix:broader_impact}
In this paper, we proposed a framework for solving GRIPs, and we demonstrated it on several key GRIP instances on several different datasets. Combining an understanding of learned regularization to solving inverse problems, together with GNNs, can have a positive potential impact on the research of GNNs as well as their applicability to solving various real-world problems. Future directions include the development of theoretical properties of GRIPs and their solutions with GNNs, in particular convergence properties and out-of-distribution behavior. We are not aware of negative societal impacts that can arise from the research presented in this paper.

\section{Architecture Details}
\label{appendix:architecture}
As discussed in Section \ref{sec:variationalMethods}, each neural regularizer (i.e., Var-GNN, ISS-GNN, and Prox-GNN) employs a GNN that is the learned graph regularizer. For example, in Var-GNN, this GNN is denoted by $\grad_{\bfx_i} \bfphi(\bfx_i, \bff_{M}, {\cal G}, \bftheta_i)$ (Equation \eqref{hyperGNN}). In ISS-GNN, the backbone GNN is denoted by $s(\bfz_{k+\hf}, \bff_{M}, {\cal G}, \bftheta_k, t_k)$ in Equation \eqref{eq:dss}. In Prox-GNN, it is denoted by $ s\left(\bfx_k - \mu \bfJ_k^{\top} (F(\bfx_k) - \bfd^{\rm obs}), \bff_{M}, {\cal G}, \bftheta_k\right)$ in Equation \eqref{eq:pgd}. 

In order to obtain a fair evaluation of the methods, we use the same GNN backbone, such that the only difference between Var-GNN, ISS-GNN, and Prox-GNN is their overall iteration definition, not the learned regularizer backbone. Specifically, we employ a simple GCN \cite{kipf2016semi} with a specified number of layers, as later discussed in Appendix \ref{appendix:hyperparameter_details}. In the case of edge recovery problem, since the property to be recovered is an edge property rather than a node one (as in all the other problems), we use a similar approach as is done with the Link Prediction task \cite{bronstein2021geometric}, and concatenate the final node features according to the given connectivity, and employ an additional MLP to predict the edge property. While other GNN backbones can be studied for specific problems, such as GAT \cite{velickovic2018graph} or GIN \cite{xu2018how}, this is not the main contribution of this paper. Instead, it is on proposing a general framework for solving GRIPs with GNNs.

\section{Datasets}
\label{appendix:datasets}
We provide information about the datasets used in our experiments, and explain why we chose them for each experiment.

\textbf{CLUSTER.}  Is a dataset introduced in \cite{dwivedi2022benchmarkinggraphneuralnetworks}, and it consists of synthetic graphs where nodes are grouped into distinct clusters, and the objective is to accurately predict the cluster membership of each node. This is a {\bf classification} problem. The CLUSTER dataset includes a total of 6 communities and, therefore, 6 classes for nodes. The graphs are of sizes $40$-$190$ nodes. The target labels for nodes are defined as the community/class labels. This dataset was used for the GRIP of graph property completion. During model training, a fixed node budget per class label per graph $nb$ is set, where $nb$ random nodes per label per graph are made visible for training with other nodes masked. During the evaluation of the test set, the same node budget $nb$ is kept for the test set. The graph property completion problem is suited for this dataset as it seeks to solve the GRIP of recovering class labels of nodes during inference on the test set, which aligns with the original intent for this dataset of node classification.
The dataset contains 10K training samples and 1K test samples.
This dataset contains no additional meta-data,i.e., no node features.

\textbf{ShapeNet.} The ShapeNet part level segmentation \cite{pytorch_geometric_shapenet} dataset offered by PyTorch Geometric, is a subset of the larger ShapeNet dataset. It includes point clouds in 16 categories with 50 classes, with each category representing a distinct type of 3D object with the class of a point being a part annotation. The dataset is commonly used for 3D object {\bf classification and part segmentation} tasks. There are $16,881$ objects in total represented by $2048$ points each, from which we sample $1024$ points uniformly.
In all experiments, graphs were generated with the K-Nearest-Neighbors algorithm with $k=10$, and using the same splits as in \citet{wang2018dynamic}. 
The associated meta-data are the normal vectors, and the 3D ($xyz$) coordinates  for respective points. The associated node states $\bfx$ are their class labels. This dataset was chosen for the graph completion problem due to the availability of meta-data, which could be utilized to aid in solving the problem and to study the effect on performance when meta-data was excluded.

\textbf{METR-LA.} The METR-LA dataset \cite{METRLA_source} consists of traffic data on a road network graph, where nodes represent traffic sensors, and edges represent the road network connectivity between the sensors. The dataset comprises 34,272 samples of the road network traffic speed at different times, at 207 nodes (sensors). The available meta-data includes the time of measurement. We follow the data splits presented in \citet{dyngrae_1}. This dataset is commonly used in {\bf regression} problems where the objective is to predict traffic at future time points. In our experiments, however, we do not perform prediction tasks and use the static graphs as-is for the inverse graph transport problem. For an example problem, consider the case where the average speed of vehicles in a road network was known for various segments, and a problem of interest was to infer the speeds of vehicles at specific points. The average speed is the sum (divided by the number of segments) of the speeds measured by sensors. This dataset fits into the framework of such a problem.  

\textbf{Chickenpox Cases in Hungary (CPOX).} This dataset was introduced in \citet{rozemberczki2021chickenpoxcaseshungarybenchmark}, and it consists of weekly chickenpox cases recorded in Hungary, represented on a graph where nodes are counties and edge weights represent the spatial proximity between the counties. There are 520 time samples, measured weekly. Usually, CPOX is used for temporal prediction tasks, where node features are lagged weekly counts of the chickenpox cases. Here we use it differently by setting lag to $0$, effectively making the target node states, $\bfx$, the number of chickenpox cases recorded in each county at the current week. This dataset was used for the inverse graph transport problem, which is a {\bf regression} task. The available meta-data is the week of measurement of chickenpox cases, and the edge weights are the distances between counties. We follow the splitting approach in  \citet{rozemberczki2021chickenpoxcaseshungarybenchmark}. The motivation behind the use of this dataset for the task of inverse source estimation is as follows: it is conceivable that it is of interest to policy makers to know the infection spread patterns across various towns (nodes), and that the available data is only the observed infection counts in towns after the spread has commenced. The GRIP would then be to infer the initial infection counts $\bfx$ of towns in order to pinpoint the source of the epidemic.

\textbf{CIFAR10 (superpixels).} This dataset was introduced in \cite{dwivedi2022benchmarkinggraphneuralnetworks}, and it includes superpixel images of the CIFAR10 dataset. Each graph includes the positions of the superpixels, their RGB value, as well as the class label of the image. To define edge weights to be recovered in the edge property recovery task, a \textbf{regression} task, we define random connectivity to the nodes of the superpixels using the Erdos-Renyi distribution with an edge probability of 5\% for each graph in a pre-processing step. The graph remains identical throughout the epochs and evaluation, such that results are deterministic. Then, we compute the edge weight to be recovered as the inverse $\ell_2$ distance of the connected nodes, according to their position in the image. The length of the history vector $\bfd^{\rm obs}$ in our experiments (results provided in Table \ref{tab:edge_rec_results}) was set to $1$ - the most difficult setting - more historical samples ($>1$) would add more information to make the problem easier. Different from the CLUSTER dataset, there is a geometrical interpretation of the superpixels with their positions and edge weights (inverse $\ell_2$ distance of the connected nodes), and so this GRIP seeks to recover the distance between the superpixels of an image.

\section{Experimental Settings}
\label{appendix:optimization_details}
The experiments were run on an Nvidia RTX A6000 GPU with $48$ GB of memory.
Each experiment was run 3 times with seeds $0,1,2$ to compute the mean and standard deviation of the performance of the different models. For GRIP experiments involving the {\bf ShapeNet} and {\bf CLUSTER} datasets, the evaluation metric was Accuracy (\%), with the data fits being assessed using the Cross-Entropy (CE) loss between the observed data $\bfd^{\rm{obs}}$ (one-hot encoded class labels) and the reconstructed data $F(F^{\dag}(\bfd^{\rm obs}))$ (the logits predicted by the models). For GRIP experiments involving the {\bf METR-LA} and {\bf CPOX} datasets, the evaluation metric was the normalized mean squared error (nMSE). We elaborate more on these for clarity in \ref{train_test_metrics}. 

The optimization for the baseline methods (LaplacianReg and TikhonovReg) were initialized with a vector of zeros, and stopped when the normalized mean squared error (nMSE) was less than $0.0025$, or when $3K$ iterations were complete. Note that the mean and standard deviation of the respective evaluation metrics for these methods are {\bf deterministic} on the test sets, and therefore have standard deviations of $0$. 

For trainable neural models, the PyTorch implementation of the Adam
optimizer is used with weight decay parameter $\bf wd$. For training, a detailed description of the hyper-parameters used can be found in \ref{appendix:hyperparameter_details}. A "{\bf max\_patience}" parameter was set such that after {\bf max\_patience} epochs if the validation loss (for regression tasks) did not improve by $0.5\%$, or the test accuracy (for classification tasks) did not improve by $1 \%$ from the best value seen thus far, the training process would terminate. Specifically, the following number of training epochs and \texttt{max\_patience}  values were set for tasks involving the following datasets: 

\begin{itemize}
    \item \textbf{METR-LA} dataset: the number of epochs was set to 100, and the \texttt{max\_patience} value was 35.
    \item \textbf{CLUSTER} dataset: the number of epochs was set to 250, and the \texttt{max\_patience} value was 50.
    \item \textbf{ShapeNet} dataset: the number of epochs was 250, with a \texttt{max\_patience} value of 3.
    \item \textbf{CPOX} dataset: the number of epochs was 250, and the \texttt{max\_patience} value was 50.
    \item \textbf{CIFAR10} dataset: the number of epochs was 250, and the \texttt{max\_patience} value was 50.
\end{itemize}

\subsection{Training and Test Loss}
\label{train_test_metrics}
In our experiments, we compute the loss for both classification and regression tasks in a normalized fashion, to be able to quantify the relative metric to ground-truth solutions. The GRIPs considered here seek to recover the node states {\bfx} which are not directly observed. For classification tasks, the total loss is the sum of the cross-entropy loss computed on the model predictions ${\bf x_{pred}}$ and the ground truth {\bfx}, and the data fit loss which is the cross-entropy between the reconstructed data $\hat{\bfd} = F(F^{\dag}(\bfd^{\rm obs}))$ and the observed data $\bfd^{\rm{obs}}$. 

For regression tasks, we employ a normalized mean squared error (nMSE) approach. The total loss is computed as the average of two normalized MSE values: one for the model predictions and one for the reconstructed data. The MSE values are normalized by the MSE of a zero vector with respect to the target data.

The normalized loss \( \mathcal{L} \) for both categories of tasks is formalized as follows:

\[
\mathcal{L} =
\begin{cases}
\text{loss}_{\bf x} + \alpha\text{loss}_{\text{data}}&,  \text{if classification} \\
\frac{1}{2} \left( \frac{\text{MSE}(\bfx, \bfx_{pred})}{\text{MSE}(0, \bfx)} + \alpha\frac{\text{MSE}({\bfd^{obs}}, {\bf \hat{d})}}{\text{MSE}(0, \bfd^{obs})} \right)&,  \text{if regression}
\end{cases}
\]
\\
where the classification losses are defined as:
\[
\text{loss}_{\bf x} = \text{CrossEntropy}(\bfx, \bfx_{pred}),
\]
$$
\text{loss}_{\text{data}} = \text{CrossEntropy}(\bfd^{obs}, {\bf \hat{d}}).
$$
The first term in the loss function for regression task problems (absent the factor of $0.5$) is the reported recovery loss nMSE on the tables, and the second term (absent the factor of $0.5$) is the reported nMSE of the data-fits. $\alpha>0$ is a hyperparameter that balances between the two terms. 

\subsection{Hyper-parameter Details}
\label{appendix:hyperparameter_details}
The hyper-parameters used in our experiments are provided for reproducibility in Tables \ref{tab:hyperparameters_cluster}, \ref{tab:hyperparameters_shapenet}, \ref{tab:hyperparameters_regression} and \ref{tab:hyperparameters_cifar10}. For the baseline Laplacian regularization experiments, the Laplacian hyper-parameter was set to be 0.1 for all experiments.

For each dataset, task, and setting, the optimal hyper-parameters were found using a Bayesian search on the most difficult setting for the given task. These optimal hyper-parameters were then used to train and test the other settings for the model for the task being considered. For example, a hyper-parameter search sweep was conducted for $nb=4$ for the CLUSTER graph completion task in Table \ref{tab:mask_results} for ISS-GNN, and the optimal hyper-parameters were used to train ISS-GNN models for $nb=8,16$. Similar sweeps were conducted for the other models. The exceptions to this were the CPOX dataset in Table \ref{tab:deblurring_results_cpox_full} for which a hyper-parameter sweep was carried out for Prox-GNN for each diffusion step setting ($k=4,8,16$), and the CIFAR10 dataset for which we used the hyper-parameters from the ShapeNet graph completion experiments. 

For reference, the hyper-parameter search space for the various hyper-parameters is given in Tables \ref{tab:hyperparameter_search_space_cluster_cpox}, \ref{tab:hyperparameter_search_space_cluster_shapenet} and \ref{tab:hyperparameter_search_space_metrla}. The search space for the batch size during training was \{8, 16, 32, 64\} for all datasets. 

We now describe the meaning of the hyper-parameters: "lr" is the learning rate of the PyTorch implementation of the Adam optimizer with "amsgrad" set to "True" and $eps=10^{-3}$,  "wd" is the weight-decay parameter of the optimizer, "layers" is the number of layers used in the network, "cglsIter" is the number of data fit iterations (Conjugate Gradient iterations) and "solveIter" is the number of model iterations, and "channels" is the length of the embedding vector for each node, i.e., number of features. 

\section{Additional Experimental Results}
\label{appendix:additional_results}
Tables \ref{tab:deblurring_results_cluster_full}, \ref{tab:pathlen_results_full}, \ref{tab:deblurring_results_cpox_full} and \ref{tab:mask_results_full} show the full range of our numerical experiments, including additional results with different settings of the GRIPs. 
\\
In Table \ref{tab:deblurring_results_cluster_full}, for the inverse source estimation problem, ISS-GNN is seen to exhibit decreasing performance with increasing diffusion steps $k$, whilst Var-GNN seems to exhibit consistent performance across $k$ values. At all diffusion steps, however, ISS-GNN and Var-GNN exhibit significant performance over Prox-GNN. All neural approaches show significant improvements from the baseline performances. 

In Table \ref{tab:pathlen_results_full} we see the results of the inverse graph transport experiments with the METR-LA dataset where increasing $pl$ makes the problem harder. For the recovery loss nMSE, an increasing trend is observed with increasing values of $pl$. All neural approaches are seen to outperform the baselines. 

Table \ref{tab:deblurring_results_cpox_full} showcases the inverse source estimation results on the CPOX dataset. All neural approaches are seen to outperform the baselines with lower nMSE of recovery for $\bfx$. The $\bfx$ recovery loss nMSE is seen to grow with increasing diffusion steps $k$. Var-GNN exhibits the best performance (lowest nMSE of recovery), followed by ISS-GNN and then Prox-GNN at all values of $k$. 

Table \ref{tab:mask_results_full} shows the results for the graph property completion experiments. For both ShapeNet and CLUSTER, an increasing node budget per label per graph $nb$ is seen to correspond to increasing accuracies for all models. Neural approaches consistently outperform the baselines. However, it is seen that for all models, a performance drop when the models were not given the available meta-data for ShapeNet is evident for all values of $nb$ (the node budget per label per graph), where the models exhibit significantly lower performance than the corresponding baseline Laplacian regularization experiments. 

Lastly, we provide an exemplary hyperparameter sensitivity study in Table \ref{tab:hyperparams_sensetivity}, starting from the best setup: cglsIter=8, channels=64, layers=16, lr=0.007, solveIter=8, wd=5e-06, and varying one parameter at a time, on the ShapeNet property completion task with nb = 4 for Var-GNN. The results demonstrate that the method is relatively stable with respect to the chosen hyperparameters.

\section{Complexity and Runtimes}
\label{appendix:complexity}
To make a quantitative comparison of the runtimes for the various trainable models, we measured the training and inference times for each. Namely, we measure the runtimes on the CLUSTER (for the graph property completion task) and CPOX (for the inverse source estimation task) for $nb=4$ and $k=4$ respectively. The GPU runtimes were measured in milliseconds per batch for training and inference. The results are provided in Table \ref{tab:timing_table}, which includes the number of trainable parameters for each model. While solutions like Var-GNN and ISS-GNN require more time and parameters than Prox-GNN, they typically offer better performance. Therefore, our framework provides several possible architectures that can be chosen depending on the needs, i.e., better downstream performance or lighter computational footprint. prox-GNN does not require many parameters by design, and so in the table, we also provide results with a variant of Prox-GNN that we name Prox-GNN (wide) such that it has a similar parameter count to Var-GNN and ISS-GNN. This was achieved by increasing the number of channels from the CPOX $k=4$ hyper-parameters for Prox-GNN in Table \ref{tab:hyperparameters_regression} from $64$ to $255$, and by increasing the number of channels from the CLUSTER $nb=4$ hyper-parameters for Prox-GNN (in Table \ref{tab:hyperparameters_cluster}) from $128$ to $183$. Note in particular that having a similar number of parameters as the other two models does not lead to Prox-GNN (wide) exhibiting comparable performance to Var-GNN and ISS-GNN.

In terms of complexity, the framework proposed here involves several iterations (denoted by 'solverIter') of the backbone GNN and data fit steps using conjugate gradient iterations which are of linear complexity (denoted by 'cglsIter'). The GNN backbone used here has a linear complexity with respect to the number of nodes and edges of the graph. Overall, the proposed networks Var-GNN, ISS-GNN, and Prox-GNN maintain the same linear space and time complexity with respect to the number of nodes $n$ and edges $m$ in the graph, similar to the backbone GNNs used in this work, such as GCN \cite{kipf2016semi}.
\clearpage
\begin{table*}[ht]
\small
\centering
\caption{Hyper-Parameter Search Space for CLUSTER (inverse source estimation) and CPOX (inverse source estimation)}
\label{tab:hyperparameter_search_space_cluster_cpox}
\begin{tabular}{|c|c|l|p{6cm}|}
\hline
\textbf{Dataset} & \textbf{Model} & \textbf{Hyper-Parameter} & \textbf{Search Space} \\ \hline

\multirow{17}{*}{\textbf{CLUSTER (inverse source estimation)}} & \multirow{6}{*}{ISS-GNN}  
& Learning Rate (lr) & \{min:0, max: 0.01\} \\ \cline{3-4}
& & Weight Decay (wd) & \{min:0, max: 0.0001\} \\ \cline{3-4}
& & Layers & \{8,16\} \\ \cline{3-4}
& & Channels & \{32,64,128\} \\ \cline{3-4}
& & cglsIter & \{8, 16, 32\} \\ \cline{3-4}
& & solveIter & \{8, 16, 32\} \\ \cline{2-4}

& \multirow{6}{*}{Var-GNN} 
& Learning Rate (lr) & \{min:0, max: 0.01\} \\ \cline{3-4}
& & Weight Decay (wd) & \{min:0, max: 0.0001\} \\ \cline{3-4}
& & Layers & \{8,16\} \\ \cline{3-4}
& & Channels & \{32,64,128\} \\ \cline{3-4}
& & cglsIter & \{8, 16, 32\} \\ \cline{3-4}
& & solveIter & \{8, 16, 32\} \\ \cline{2-4}

& \multirow{6}{*}{Prox-GNN} 
& Learning Rate (lr) & \{min:0, max: 0.01\} \\ \cline{3-4}
& & Weight Decay (wd) & \{min:0, max: 0.0001\} \\ \cline{3-4}
& & Layers & \{8,16\} \\ \cline{3-4}
& & Channels & \{32,64,128\} \\ \cline{3-4}
& & cglsIter & \{8, 16, 32\} \\ \cline{3-4}
& & solveIter & \{8, 16, 32\} \\ \hline

\multirow{17}{*}{\textbf{CPOX (inverse source estimation)}} & \multirow{6}{*}{ISS-GNN}  
& Learning Rate (lr) & \{min:0, max: 0.01\} \\ \cline{3-4}
& & Weight Decay (wd) & \{min:0, max: 0.0001\} \\ \cline{3-4}
& & Layers & \{8, 16, 32\} \\ \cline{3-4}
& & Channels & \{32, 64, 128\} \\ \cline{3-4}
& & cglsIter & \{4, 8, 16\} \\ \cline{3-4}
& & solveIter & \{16, 32, 50\} \\ \cline{2-4}

& \multirow{6}{*}{Var-GNN} 
& Learning Rate (lr) & \{min:0, max: 0.01\} \\ \cline{3-4}
& & Weight Decay (wd) & \{min:0, max: 0.0001\} \\ \cline{3-4}
& & Layers & \{8, 16, 32, 64\} \\ \cline{3-4}
& & Channels & \{32, 64, 128\} \\ \cline{3-4}
& & cglsIter & \{8, 16, 32\} \\ \cline{3-4}
& & solveIter & \{8, 16, 32\} \\ \cline{2-4}

& \multirow{6}{*}{Prox-GNN} 
& Learning Rate (lr) & \{min:0, max: 0.01\} \\ \cline{3-4}
& & Weight Decay (wd) & \{min:0, max: 0.0001\} \\ \cline{3-4}
& & Layers & \{8,16,32\} \\ \cline{3-4}
& & Channels & \{32,64,128\} \\ \cline{3-4}
& & cglsIter & \{5, 8, 10\} \\ \cline{3-4}
& & solveIter & \{15, 20, 30, 40\} \\ \hline

\end{tabular}
\end{table*}
\clearpage
\begin{table*}[ht]
\small
\centering
\caption{Hyper-Parameter Search Space for CLUSTER (property completion) and ShapeNet (property completion)}
\label{tab:hyperparameter_search_space_cluster_shapenet}
\begin{tabular}{|c|c|l|p{6cm}|}
\hline
\textbf{Dataset} & \textbf{Model} & \textbf{Hyper-Parameter} & \textbf{Search Space} \\ \hline

\multirow{17}{*}{\textbf{CLUSTER (property completion)}} & \multirow{6}{*}{ISS-GNN}  
& Learning Rate (lr) & \{min:0, max: 0.01\} \\ \cline{3-4}
& & Weight Decay (wd) & \{min:0, max: 0.0001\} \\ \cline{3-4}
& & Layers & \{8,16\} \\ \cline{3-4}
& & Channels & \{32,64,128\} \\ \cline{3-4}
& & cglsIter & \{8, 16, 32\} \\ \cline{3-4}
& & solveIter & \{8, 16, 32\} \\ \cline{2-4}

& \multirow{6}{*}{Var-GNN} 
& Learning Rate (lr) & \{min:0, max: 0.01\} \\ \cline{3-4}
& & Weight Decay (wd) & \{min:0, max: 0.0001\} \\ \cline{3-4}
& & Layers & \{8, 16\} \\ \cline{3-4}
& & Channels & \{32, 64, 128\} \\ \cline{3-4}
& & cglsIter & \{8, 16, 32\} \\ \cline{3-4}
& & solveIter & \{8, 16, 32\} \\ \cline{2-4}

& \multirow{6}{*}{Prox-GNN} 
& Learning Rate (lr) & \{min:0, max: 0.01\} \\ \cline{3-4}
& & Weight Decay (wd) & \{min:0, max: 0.0001\} \\ \cline{3-4}
& & Layers & \{8, 16\} \\ \cline{3-4}
& & Channels & \{32,64,128\} \\ \cline{3-4}
& & cglsIter & \{8, 16, 32\} \\ \cline{3-4}
& & solveIter & \{8, 16, 32\} \\ \hline

\multirow{17}{*}{\textbf{ShapeNet (property completion)}} & \multirow{6}{*}{ISS-GNN}  
& Learning Rate (lr) & \{min:0, max: 0.01\} \\ \cline{3-4}
& & Weight Decay (wd) & \{min:0, max: 0.0001\} \\ \cline{3-4}
& & Layers & \{8, 16\} \\ \cline{3-4}
& & Channels & \{32, 64, 80\} \\ \cline{3-4}
& & cglsIter & \{8, 16, 32\} \\ \cline{3-4}
& & solveIter & \{8, 16, 32\} \\ \cline{2-4}

& \multirow{6}{*}{Var-GNN} 
& Learning Rate (lr) & \{min:0, max: 0.01\} \\ \cline{3-4}
& & Weight Decay (wd) & \{min:0, max: 0.0001\} \\ \cline{3-4}
& & Layers & \{8, 16, 32\} \\ \cline{3-4}
& & Channels & \{32, 64, 128\} \\ \cline{3-4}
& & cglsIter & \{8, 16, 32\} \\ \cline{3-4}
& & solveIter & \{8, 16, 32\} \\ \cline{2-4}

& \multirow{6}{*}{Prox-GNN} 
& Learning Rate (lr) & \{min:0, max: 0.01\} \\ \cline{3-4}
& & Weight Decay (wd) & \{min:0, max: 0.0001\} \\ \cline{3-4}
& & Layers & \{8,16\} \\ \cline{3-4}
& & Channels & \{32,64,128\} \\ \cline{3-4}
& & cglsIter & \{8, 16, 32\} \\ \cline{3-4}
& & solveIter & \{8, 16, 32\} \\ \hline

\end{tabular}
\end{table*}
\clearpage
\clearpage
\begin{table*}[ht]
\small
\centering
\caption{Hyper-Parameter Search Space for METR-LA (inverse transport)}
\label{tab:hyperparameter_search_space_metrla}
\begin{tabular}{|c|c|l|p{6cm}|}
\hline
\textbf{Dataset} & \textbf{Model} & \textbf{Hyper-Parameter} & \textbf{Search Space} \\ \hline

\multirow{17}{*}{\textbf{METR-LA (inverse transport)}} & \multirow{6}{*}{ISS-GNN}  
& Learning Rate (lr) & \{min:0, max: 0.01\} \\ \cline{3-4}
& & Weight Decay (wd) & \{min:0, max: 0.0001\} \\ \cline{3-4}
& & Layers & \{8,16,32,64\} \\ \cline{3-4}
& & Channels & \{32, 64, 128\} \\ \cline{3-4}
& & cglsIter & \{8, 16, 32\} \\ \cline{3-4}
& & solveIter & \{8, 16, 32\} \\ \cline{2-4}

& \multirow{6}{*}{Var-GNN} 
& Learning Rate (lr) & \{min:0, max: 0.01\} \\ \cline{3-4}
& & Weight Decay (wd) & \{min:0, max: 0.0001\} \\ \cline{3-4}
& & Layers & \{8, 16, 32, 64\} \\ \cline{3-4}
& & Channels & \{32, 64, 128\} \\ \cline{3-4}
& & cglsIter & \{8, 16, 32\} \\ \cline{3-4}
& & solveIter & \{8, 16, 32\} \\ \cline{2-4}

& \multirow{6}{*}{Prox-GNN} 
& Learning Rate (lr) & \{min:0, max: 0.01\} \\ \cline{3-4}
& & Weight Decay (wd) & \{min:0, max: 0.0001\} \\ \cline{3-4}
& & Layers & \{8, 16, 32, 64\} \\ \cline{3-4}
& & Channels & \{32,64,128\} \\ \cline{3-4}
& & cglsIter & \{8, 16, 32\} \\ \cline{3-4}
& & solveIter & \{8, 16, 32\} \\ \hline

\end{tabular}
\end{table*}

\clearpage

\begin{table*}[ht]
\small
\centering
\caption{The hyper-parameters for CLUSTER (inverse source estimation) and CLUSTER (property completion).}
\label{tab:hyperparameters_cluster}
\begin{tabular}{|c|l|l|p{6cm}|}
\hline
\textbf{Dataset (task)} & \textbf{Method} & \textbf{Setting} & \textbf{Hyper-parameters} \\ \hline

\multirow{15}{*}{\textbf{CLUSTER (inv. source estimation)}} & \multirow{3}{*}{ISS-GNN}  
& $k$ = 4 & batch\_size = 8, lr = 0.00583, wd = 3.43e-05, layers = 8, channels = 32, cglsIter = 32, solveIter = 16 \\ \cline{3-4}
& & $k$ = 8 & batch\_size=8, cglsIter=32, channels=32, layers=8, lr=0.0058, solveIter=16, wd=3.43e-05 \\ \cline{3-4}
& & $k$ = 16 & batch\_size=8 , cglsIter=32, channels=32, layers=8, lr=0.0058, solveIter=16, wd=3.43e-05\\ \cline{2-4}

& \multirow{3}{*}{Var-GNN} 
& $k$ = 4 & batch\_size=8, cglsIter=8, channels=64, layers=8, lr=0.0018, solveIter=16, wd=3.81e-06\\ \cline{3-4}
& & $k$ = 8 & batch\_size=8, cglsIter=8, channels=64, layers 8, lr=0.0018, solveIter=16, wd=3.81e-06 \\ \cline{3-4}
& & $k$ = 16 & batch\_size=8, cglsIter=8, channels=64, layers 8, lr=0.0018, solveIter=16, wd=3.81e-06
 \\ \cline{2-4}

& \multirow{3}{*}{Prox-GNN} 
& $k$ = 4 & batch\_size=8, cglsIter=8, channels=128, layers 8, lr=0.0058, solveIter=16, wd=3.81e-06
 \\ \cline{3-4}
& & $k$ = 8 & batch\_size=8, cglsIter=8, channels=128, layers=8, lr=0.0058, solveIter=16, wd=3.81e-06
 \\ \cline{3-4}
& & $k$ = 16 & batch\_size=8, cglsIter=8, channels=128, layers=8, lr=0.0058, solveIter=16, wd=3.81e-06
 \\ \cline{2-4}

& \multirow{3}{*}{LapReg} 
& $k$ = 4 & batch\_size=1, step\_size=0.1, solveIter=3000 \\ \cline{3-4}
& & $k$ = 8 & batch\_size=1, step\_size=0.1, solveIter=3000 \\ \cline{3-4}
& & $k$ = 16 & batch\_size=1, step\_size=0.1, solveIter=3000 \\ \cline{2-4}

& \multirow{3}{*}{TikhonovReg} 
& $k$ = 4 & batch\_size=1, step\_size=0.1, solveIter=3000 \\ \cline{3-4}
& & $k$ = 8 & batch\_size=1, step\_size=0.1, solveIter=3000 \\ \cline{3-4}
& & $k$ = 16 & batch\_size=1, step\_size=0.1, solveIter=3000 \\ \hline

\multirow{15}{*}{\textbf{CLUSTER (prop. completion)}} & \multirow{3}{*}{ISS-GNN}  
& nb = 4 & batch\_size=64, cglsIter=32, channels=32, layers=8, lr=0.0058, solveIter=16, wd=3.43e-05 \\ \cline{3-4}
& & nb = 8 & batch\_size=64, cglsIter=32, channels=32, layers=8, lr=0.0058, solveIter=16, wd=3.43e-05 \\ \cline{3-4}
& & nb = 16 & batch\_size=64, cglsIter=32, channels=32, layers=8, lr=0.0058, solveIter=16, wd=3.43e-05 \\ \cline{2-4}

& \multirow{3}{*}{Var-GNN} 
& nb = 4 & batch\_size=32, cglsIter=8, channels=64, layers=8, lr=0.0018, solveIter=16, wd=3.81e-06 \\ \cline{3-4}
& & nb = 8 & batch\_size=32, cglsIter=8, channels=64, layers=8, lr=0.0018, solveIter=16, wd=3.81e-06 \\ \cline{3-4}
& & nb = 16 & batch\_size=32, cglsIter=8, channels=64, layers=8, lr=0.0018, solveIter=16, wd=3.81e-06 \\ \cline{2-4}

& \multirow{3}{*}{Prox-GNN} 
& nb = 4 & batch\_size=32, cglsIter=8, channels=128, layers=8, lr=0.0058, solveIter=16, wd=3.81e-06 \\ \cline{3-4}
& & nb = 8 & batch\_size=32, cglsIter=8, channels=128, layers=8, lr=0.0058, solveIter=16, wd=3.81e-06 \\ \cline{3-4}
& & nb = 16 & batch\_size=32, cglsIter=8, channels=128, layers=8, lr=0.0058, solveIter=16, wd=3.81e-06 \\ \cline{2-4}

& \multirow{3}{*}{LapReg} 
& nb = 4 & batch\_size=1, step\_size=0.1, solveIter=3000 \\ \cline{3-4}
& & nb = 8 & batch\_size=1, step\_size=0.1, solveIter=3000 \\ \cline{3-4}
& & nb = 16 & batch\_size=1, step\_size=0.1, solveIter=3000 \\ \cline{2-4}

& \multirow{3}{*}{TikhonovReg} 
& nb = 4 & batch\_size=1, step\_size=0.1, solveIter=3000 \\ \cline{3-4}
& & nb = 8 & batch\_size=1, step\_size=0.1, solveIter=3000 \\ \cline{3-4}
& & nb = 16 & batch\_size=1, step\_size=0.1, solveIter=3000 \\ \hline

\end{tabular}
\end{table*}
\clearpage
\begin{table*}[ht]
\small
\centering
\caption{The hyper-parameters for ShapeNet (property completion).}
\label{tab:hyperparameters_shapenet}
\begin{tabular}{|c|l|l|p{6cm}|}
\hline
\textbf{Dataset (task)} & \textbf{Method} & \textbf{Setting} & \textbf{Hyper-parameters} \\ \hline

\multirow{15}{*}{\textbf{ShapeNet (prop. completion)}} & \multirow{3}{*}{ISS-GNN}  
& nb = 4 & batch\_size=14, cglsIter=32, channels=32, layers=16, lr=0.00933, solveIter=16, wd=1.11489e-06 \\ \cline{3-4}
& & nb = 8 & batch\_size=14, cglsIter=32, channels=32, layers=16, lr=0.0093, solveIter=16, wd=1.1149e-06 \\ \cline{3-4}
& & nb = 16 & batch\_size=14, cglsIter=32, channels=32, layers=16, lr=0.00933, solveIter=16, wd=1.11489e-06 \\ \cline{2-4}

& \multirow{3}{*}{Var-GNN} 
& nb = 4 & batch\_size=14, cglsIter=8, channels=64, layers=16, lr=0.00758, solveIter=8, wd=4.79e-06 \\ \cline{3-4}
& & nb = 8 & batch\_size=14, cglsIter=8, channels=64, layers=16, lr=0.00758, solveIter=8, wd=4.79e-06 \\ \cline{3-4}
& & nb = 16 & batch\_size=14, cglsIter=8, channels=64, layers=16, lr=0.00758, solveIter=8, wd=4.79e-06 \\ \cline{2-4}

& \multirow{3}{*}{Prox-GNN} 
& nb = 4 & batch\_size=14, cglsIter=16, channels=128, layers=16, lr=0.00343, solveIter=8, wd=1.95e-05
 \\ \cline{3-4}
& & nb = 8 & batch\_size=14, cglsIter=16, channels=128, layers=16, lr=0.00343, solveIter=8, wd=1.95e-05
 \\ \cline{3-4}
& & nb = 16 & batch\_size=14, cglsIter=16, channels=128, layers=16, lr=0.00343, solveIter=8, wd=1.95e-05
 \\ \cline{2-4}

& \multirow{3}{*}{LapReg} 
& nb = 4 & batch\_size=1, step\_size=0.02, solveIter=3000 \\ \cline{3-4}
& & nb = 8 & batch\_size=1, step\_size=0.02, solveIter=3000  \\ \cline{3-4}
& & nb = 16 & batch\_size=1, step\_size=0.02, solveIter=3000 \\ \cline{2-4}

& \multirow{3}{*}{TikhonovReg} 
& nb = 4 &  batch\_size=1, step\_size=0.02, solveIter=3000  \\ \cline{3-4}
& & nb = 8 &  batch\_size=1, step\_size=0.02, solveIter=3000  \\ \cline{3-4}
& & nb = 16 &  batch\_size=1, step\_size=0.02, solveIter=3000  \\ \hline

\end{tabular}
\end{table*}
\clearpage
\begin{table*}[ht]
\small
\centering
\caption{The hyper-parameters for METR-LA (inverse graph transport) and CPOX (inverse source estimation). The number of diffusion steps is denoted by $k$ for CPOX. The path length is denoted by $pl$ for METR-LA.}
\label{tab:hyperparameters_regression}
\begin{tabular}{|c|l|l|p{8cm}|}
\hline
\textbf{Dataset (task)} & \textbf{Method} & \textbf{Setting} & \textbf{Hyper-parameters} \\ \hline

\multirow{15}{*}{\textbf{CPOX (inv. source estimation)}} & \multirow{3}{*}{ISS-GNN}  
& $k$ = 4 & batch\_size=64, cglsIter=16, channels=64, layers=16, lr=0.00899, solveIter=32, wd=9.75e-05
 \\ \cline{3-4}
& & $k$ = 8 & batch\_size=64, cglsIter=16, channels=64, layers=16, lr=0.00899, solveIter=32, wd=9.75e-05
 \\ \cline{3-4}
& & $k$ = 16 & batch\_size=64, cglsIter=16, channels=64, layers=16, lr=0.00899, solveIter=32, wd=9.75e-05
 \\ \cline{2-4}

& \multirow{3}{*}{Var-GNN} 
& $k$ = 4 & batch\_size=64, cglsIter=32, channels=32, layers=8, lr=0.00028, solveIter=8, wd=7.77e-05
 \\ \cline{3-4}
& & $k$ = 8 & batch\_size=64, cglsIter=32, channels=32, layers=8, lr=0.00028, solveIter=8, wd=7.77e-05
 \\ \cline{3-4}
& & $k$ = 16 & batch\_size=64, cglsIter=32, channels=32, layers=8, lr=0.00028, solveIter=8, wd=7.77e-05

 \\ \cline{2-4}

& \multirow{3}{*}{Prox-GNN} 
& $k$ = 4 & batch\_size=64, cglsIter=5, channels=64, layers=8, lr=0.00680, solveIter=40, wd=4.08e-05 \\ \cline{3-4}
& & $k$ = 8 & batch\_size=64, cglsIter=10, channels=64, layers=16, lr=0.00785, solveIter=40, wd=9.28e-05
 \\ \cline{3-4}
& & $k$ = 16 & batch\_size=64, cglsIter=10, channels=32, layers=32, lr=0.00849, solveIter=40, wd=1.07e-05
 \\ \cline{2-4}

& \multirow{3}{*}{LapReg} 
& $k$ = 4 & batch\_size=1, step\_size=0.00005, solveIter=3000
 \\ \cline{3-4}
& & $k$ = 8 & batch\_size=1, step\_size=0.00003, solveIter=3000
 \\ \cline{3-4}
& & $k$ = 16 & batch\_size=1, step\_size=0.00003, solveIter=3000
 \\ \cline{2-4}

& \multirow{3}{*}{TikhonovReg} 
&  $k$ = 4 & batch\_size=1, step\_size=0.0002, solveIter=3000
 \\ \cline{3-4}
& & $k$ = 8 & batch\_size=1, step\_size=0.0002, solveIter=3000
 \\ \cline{3-4}
& & $k$ = 16 & batch\_size=1, step\_size=0.0002, solveIter=3000
 \\ \hline

\multirow{15}{*}{\textbf{METR-LA (inv. graph transport)}} & \multirow{3}{*}{ISS-GNN}  
& $pl$ = 8 & batch\_size=8, cglsIter=8, channels=128, layers=8, lr=0.00309, solveIter=32, wd=5.93e-05 \\ \cline{3-4}
& & $pl$ = 16 & batch\_size=8, cglsIter=8, channels=128, layers=8, lr=0.00309, solveIter=32, wd=5.93e-05
 \\ \cline{3-4}
& & $pl$ = 32 & batch\_size=8, cglsIter=8, channels=128, layers=8, lr=0.00309, solveIter=32, wd=5.93e-05 \\ \cline{2-4}

& \multirow{3}{*}{Var-GNN} 
& $pl$ = 8 & batch\_size=128, cglsIter=8, channels=32, layers=8, lr=0.00250, solveIter=8, wd=9.89e-06  \\ \cline{3-4}
& & $pl$ = 16 & batch\_size=128, cglsIter=8, channels=32, layers=8, lr=0.00250, solveIter=8, wd=9.89e-06 \\ \cline{3-4}
& & $pl$ = 32 & batch\_size=128, cglsIter=8, channels=32, layers=8, lr=0.00250, solveIter=8, wd=9.89e-06 \\ \cline{2-4}

& \multirow{3}{*}{Prox-GNN} 
& $pl$ = 8 & batch\_size=32, cglsIter=8, channels=128, layers=8, lr=0.00309, solveIter=32, wd=5.93e-05
 \\ \cline{3-4}
& & $pl$ = 16 & batch\_size=32, cglsIter=8, channels=128, layers=8, lr=0.00309, solveIter=32, wd=5.93e-05
 \\ \cline{3-4}
& & $pl$ = 32 & batch\_size=32, cglsIter=8, channels=128, layers=8, lr=0.00309, solveIter=32, wd=5.93e-05
 \\ \cline{2-4}

& \multirow{3}{*}{LapReg} 
& $pl$ = 8 & batch\_size=1, step\_size=0.001, solveIter=3000 \\ \cline{3-4}
& & $pl$ = 16 & batch\_size=1, step\_size=0.0005, solveIter=3000
 \\ \cline{3-4}
& & $pl$ = 32 & batch\_size=1, step\_size=0.0005, solveIter=3000 \\ \cline{2-4}

& \multirow{3}{*}{TikhonovReg} 
& $pl$ = 8 & batch\_size=1, step\_size=0.001, solveIter=3000 \\ \cline{3-4}
& & $pl$ = 16 & batch\_size=1, step\_size=0.001, solveIter=3000 \\ \cline{3-4}
& & $pl$ = 32 & batch\_size=1, step\_size=0.001, solveIter=3000 \\ \hline

\end{tabular}
\end{table*}

\begin{table*}[ht]
\small
\centering
\caption{The hyper-parameters for CIFAR-10 dataset for the Edge Recovery Problem. }
\label{tab:hyperparameters_cifar10}
\begin{tabular}{|c|l|p{10cm}|}
\hline
\textbf{Dataset} & \textbf{Method} & \textbf{Hyper-parameters} \\ \hline

\multirow{6}{*}{\textbf{CIFAR-10}} & ISS-GNN  
& batch\_size=14, cglsIter=32, channels=32, layers=16, lr=0.00933, solveIter=16, wd=1.11489e-06 \\ \cline{2-3}

& Var-GNN  
&  batch\_size=14, cglsIter=8, channels=64, layers=16, lr=0.00758, solveIter=8, wd=4.79e-06 \\ \cline{2-3}

& Prox-GNN 
& batch\_size=14, cglsIter=16, channels=128, layers=16, lr=0.00343, solveIter=8, wd=1.95e-05 \\ \cline{2-3}

& LapReg  
& batch\_size=1, step\_size=0.02, solveIter=3000 \\ \cline{2-3}

& TikhonovReg  
& batch\_size=1, step\_size=0.02, solveIter=3000 \\ \hline

\end{tabular}
\end{table*}

\begin{table*}[t]
    \centering
    \caption{Training and inference GPU runtimes (milliseconds) per batch, number of parameters (thousands), and node classification accuracy (\%) on CLUSTER ($nb=4$), and Mean Squared Error (MSE) on CPOX ($k=4$ diffusion steps). Hyper-parameters for each model and task are as given in Table \ref{tab:hyperparameters_cluster} and Table \ref{tab:hyperparameters_regression}. Times are measured on a 48 GB Nvidia RTX A6000 GPU. Because Prox-GNN does not require many parameters by design, we also provide results with a variant with enlarged hidden dimension called Prox-GNN (wide), such that it has a similar parameter count to Var-GNN and ISS-GNN. }
    \label{tab:timing_table}
    \resizebox{\textwidth}{!}{
    \begin{tabular}{lcccccccccc}
        \toprule
        & \multicolumn{4}{c}{\textbf{CLUSTER (graph prop. completion), $nb=4$}} & & \multicolumn{4}{c}{\textbf{CPOX (inv. source estimation), $k=4$}} \\
        \cmidrule(lr){2-5} \cmidrule(lr){7-10}
        \textbf{Metric} & \textbf{Var-GNN} & \textbf{ISS-GNN} & \textbf{Prox-GNN} & \textbf{Prox-GNN (wide)} & & \textbf{Var-GNN} & \textbf{ISS-GNN} & \textbf{Prox-GNN} & \textbf{Prox-GNN (wide)} \\
        \midrule
        Training time (ms) & $326.56 \pm 14.36$ & $337.99 \pm 20.32$ & $110.90 \pm 11.81$ & $104.00 \pm 7.34$ & & $383.76 \pm 32.72$ & $968.47 \pm 66.41$ & $250.75 \pm 19.24$ & $237.77 \pm 13.03$ \\
        Inference time (ms) & $152.27 \pm 40.17$ & $150.01 \pm 14.47$ & $43.65 \pm 6.55$ & $39.09 \pm 5.35$ & & $178.17 \pm 0.36$ & $416.51 \pm 36.56$ & $84.63 \pm 10.63$ & $79.39 \pm 0.85$ \\
        Parameters (K) & $112.710$ & $28.710$ & $58.151$ & $113.041$ & & $28.353$ & $198.913$ & $13.218$ & $198.679$ \\
        Accuracy (\%) & $88.89 \pm 0.51$ & $57.08 \pm 7.36$ & $53.58 \pm 2.58$ & $53.67 \pm 2.65$ & & N/A & N/A & N/A & N/A \\
        nMSE(\bfx) & N/A & N/A & N/A & N/A & & $0.29 \pm 0.001$ & $0.39 \pm 1.4\cdot 10^{-3}$ & $0.57 \pm 0$ & $0.57 \pm 0$ \\
        \bottomrule
    \end{tabular}
    }
\end{table*}

\begin{table*}[t]
\centering
\caption{Inverse Source Estimation Test Accuracy (\%, $\uparrow$) on CLUSTER Dataset, with $k=4,8,16$ diffusion steps. Shown is the mean $\pm$ std for 3 runs with different seeds.}
\label{tab:deblurring_results_cluster_full}
\begin{tabular}{lcccccc}
\toprule
 Method & \multicolumn{2}{c}{$k=4$} & \multicolumn{2}{c}{$k=8$} & \multicolumn{2}{c}{$k=16$} \\
\cmidrule(r){2-3} \cmidrule(r){4-5} \cmidrule(r){6-7}
& Accuracy & CE(data) & Accuracy & CE(data) & Accuracy & CE(data) \\
\midrule
 LapReg & $26.30$ & $1.7$ & $26.28$ & $1.78$ & $26.28 $ & $1.75$ \\
 TikhonovReg & $26.34$ & $1.78$ & $26.28$ & $1.78$ & $26.28$ & $1.76$ \\
 \midrule
 Var-GNN & $79.61 \pm 1.40$ & $1.75 \pm 7.61 \cdot 10^{-5}$ & $82.73 \pm 0.88$ & $1.75 \pm 7.17 \cdot 10^{-8}$ & $82.36 \pm 0.20$ & $1.75 \pm 1.8 \cdot 10^{-9}$ \\
  ISS-GNN & $87.11 \pm 2.95$ & $1.75 \pm 4.13 \cdot 10^{-7} $ & $66.87 \pm 2.41$ & $1.75 \pm 4.82 \cdot 10^{-8}$ & $52.31 \pm 4.18$ &  $1.75 \pm 3.1 \cdot 10^{-9}$ \\
  Prox-GNN & $43.35 \pm 2.12$ & $1.75 \pm 8.1 \cdot 10^{-6}$ & $43.78 \pm 3.16$ & $1.75 \pm 8.25 \cdot 10^{-9}$ & $44.48 \pm 3.63$ & $1.75 \pm 3.8 \cdot 10^{-9}$ \\
\bottomrule
\end{tabular}%
\end{table*}
 
\begin{table*}[t]
\small
\centering
\caption{Inverse Graph Transport normalized-mean-squared-error (nMSE) on the METR-LA dataset, showing the nMSE of the recovery of $\bfx$ and the nMSE of the data fit.}
\label{tab:pathlen_results_full}
\resizebox{\textwidth}{!}{
\begin{tabular}{lcccccc}
\toprule
 Model & \multicolumn{2}{c}{$pl=8$} & \multicolumn{2}{c}{$pl=16$} & \multicolumn{2}{c}{$pl=32$} \\
\cmidrule(r){2-3} \cmidrule(r){4-5} \cmidrule(r){6-7}
 & nMSE $\downarrow$ & Data Fit (nMSE $\downarrow$) & nMSE $\downarrow$ & Data Fit (nMSE $\downarrow$) & nMSE $\downarrow$ & Data Fit (nMSE $\downarrow$) \\
\midrule
   LaplacianReg & $0.43$ & $0.10$ & $0.56$ & $0.13$  & $0.61$& $0.11$ \\
 TikhonovReg & $0.25$ & $0.04$ & $0.31$ & $0.03$ & $0.36$ & $0.04$ \\
 \midrule
  Var-GNN & $0.003 \pm 5.6 \cdot 10^{-5}$ & $4.7 \cdot 10^{-6} \pm 1.4 \cdot 10^{-7}$ & $0.004 \pm 8.6 \cdot 10^{-5}$ & $7.2 \cdot 10^{-6} \pm  5.2 \cdot 10^{-7}$ & $0.004 \pm 2.0 \cdot 10^{-5}$ & $8.2 \cdot 10^{-6} \pm 0.00$ \\
   ISS-GNN & $0.14 \pm 2.5 \cdot 10^{-4}$ & $0.003 \pm 5.9 \cdot 10^{-5}$ & $0.17 \pm 1.1 \cdot 10^{-3}$ & $0.004 \pm 6.9 \cdot 10^{-5}$ & $0.21 \pm 5.9 \cdot 10^{-5}$ & $0.004 \pm 7.9 \cdot 10^{-6}$ \\
  Prox-GNN & $0.10 \pm 2.0 \cdot 10^{-3}$ & $0.001 \pm 7.8 \cdot 10^{-5}$ & $0.14 \pm 1.1 \cdot 10^{-3}$ & $0.002 \pm 6.4 \cdot 10^{-5}$ & $0.21\pm 3.5 \cdot 10^{-3}$ & $ 0.003 \pm 1.4 \cdot 10^{-4}$ \\
\bottomrule
\end{tabular}%
}
\end{table*}

\begin{table*}[t]
\small
\centering
\caption{Inverse Source Estimation normalized-mean-squared error (nMSE) of the recovery of $\bfx$ and the data-fit on the Chickenpox-Hungary dataset.}
\label{tab:deblurring_results_cpox_full}
\resizebox{\textwidth}{!}{%
\begin{tabular}{lcccccc}
\toprule
 Method & \multicolumn{2}{c}{$k=4$} & \multicolumn{2}{c}{$k=8$} & \multicolumn{2}{c}{$k=16$} \\
\cmidrule(r){2-3} \cmidrule(r){4-5} \cmidrule(r){6-7}
&  nMSE $\downarrow$ & Data Fit (nMSE $\downarrow$) & nMSE $\downarrow$ & Data Fit (nMSE $\downarrow$) & nMSE $\downarrow$ & Data Fit (nMSE $\downarrow$) \\
\midrule
  LaplacianReg& $0.78$ & $0.082$ & $0.83$& $0.068$ & $0.84$ & $0.048$ \\
 TikhonovReg & $0.74$ & $0.075$ & $0.81$ & $0.059$ & $0.84$ & $0.051$ \\
 \midrule
  Var-GNN & $0.29 \pm 0.001$ & $1.01 \cdot 10^{-7} \pm  6.70 \cdot 10^{-9}$ & $0.41 \pm 7.5 \cdot 10^{-4}$ & $9.05 \cdot 10^{-11} \pm 1.89\cdot 10^{-11}$ & $0.59 \pm 0.01$ & $ 5.9 \cdot 10^{-11} \pm 0.00$ \\
   ISS-GNN & $0.39 \pm 1.4\cdot 10^{-3}$ & $1.34 \cdot 10^{-6} \pm 1.88 \cdot 10^{-8}$ & $0.53 \pm 2.2 \cdot 10^{-3}$ & $8.86 \cdot 10^{-10} \pm 3.98 \cdot 10^{-12}$ & $0.59 \pm 8.4\cdot 10^{-5}$ & $5.5 \cdot 10^{-11} \pm 0.00$ \\
  Prox-GNN & $0.57 \pm 0$ & $ 5.7 \cdot 10^{-4} \pm 0$ & $0.66 \pm 0$ & $7.20 \cdot 10^{-4} \pm 0$ & $0.73 \pm 0.00$ & $1\cdot 10^{-5} \pm 0$ \\
\bottomrule
\end{tabular}}
\end{table*}

\begin{table*}[t]
\footnotesize
\centering
\caption{Graph property completion performance (accuracy$\pm$standard deviation (\%)) on the ShapeNet and CLUSTER Datasets. The node budget per label per graph is denoted by '$nb$'. An asterisk (*) indicates no meta-data was used.}
\label{tab:mask_results_full}
\resizebox{\textwidth}{!}{%
\begin{tabular}{llcccccc}
\toprule
Dataset & Model & \multicolumn{2}{c}{$nb=4$} & \multicolumn{2}{c}{$nb=8$} & \multicolumn{2}{c}{$nb=16$} \\
\cmidrule(r){3-4} \cmidrule(r){5-6} \cmidrule(r){7-8}
& & Accuracy (\%)$\uparrow$ & Data Fit (CE $\downarrow$) & Accuracy (\%)$\uparrow$ & Data Fit (CE $\downarrow$) & Accuracy (\%)$\uparrow$ & Data Fit (CE $\downarrow$) \\
\midrule
\multirow{9}{*}{ShapeNet} &
 LaplacianReg & $74.90$ & $3.87$ & $79.02$ & $3.88$ & $82.08$ & $3.88$ \\
 & TikhonovReg & $6.91$ & $3.71$ & $7.979$ & $3.81$ & $10.07$ & $3.85$ \\
\cmidrule(r){2-8}
 & Var-GNN & $89.16 \pm 0.31$ & $0.24 \pm 0.08$ & $90.33 \pm 0.054$ & $0.15 \pm 0.06$ & $92.05 \pm 0.31$ & $0.13 \pm 0.014$ \\
 & Var-GNN* & $26.89 \pm 0.02$ & $0.37 \pm 0.02$ & $33.03 \pm 0.057$ & $0.24 \pm 0.088$ & $43.80 \pm 0.042 $ & $0.23 \pm 0.059$ \\
 &ISS-GNN & $89.59 \pm 0.53$ & $0.27 \pm 0.02$& $90.26 \pm 0.45$ & $0.27 \pm 0.043$ & $91.94 \pm 0.49 $ & $0.23 \pm 0.051$ \\
& ISS-GNN* & $26.45 \pm 0.02$ & $3.01 \pm 1.0 \cdot 10^{-6}$ & $31.72 \pm 0.88$ & $2.50 \pm 0.38$ & $42.96 \pm 0.14 $ & $2.97 \pm 0.015$ \\
 & Prox-GNN & $89.33 \pm 0.23$ & $2.94 \pm 2.8 \cdot 10^{-9}$ & $90.57 \pm 0.28$ & $2.94 \pm 9.14 \cdot 10^{-9}$ & $92.40 \pm 0.31$ & $2.95 \pm 4.63 \cdot 10^{-9}$ \\
 & Prox-GNN* & $26.96 \pm 0.01$ & $2.94 \pm 4.1 \cdot 10^{-9}$ & $33.22 \pm 6.7 \cdot 10^{-3}$ & $2.94 \pm 1.09 \cdot 10^{-8}$ & $44.01 \pm 0.018$ & $2.94 \pm 8.32 \cdot 10^{-9}$ \\
\midrule
\multirow{6}{*}{CLUSTER} & 
 LaplacianReg & $70.13$ & $1.78$ & $83.27$ & $1.78$ & $76.64$ & $1.78$ \\
  & TikhonovReg & $34.64$ & $1.12$ & $51.40$ & $1.12$ & $77.09$ & $1.24$ \\
\cmidrule(r){2-8}
 & Var-GNN & $88.89 \pm 0.51$ & $1.04 \pm 4.9 \cdot 10^{-9}$ & $93.18\pm 0.21$ & $1.043 \pm 1.41 \cdot 10^{-8}$ & $97.95 \pm 0.16$ & $1.04 \pm 1.05 \cdot 10^{-8}$ \\
&ISS-GNN & $57.08 \pm 7.36$ & $1.04 \pm 1.6 \cdot 10^{-8}$ & $67.13 \pm 3.55$ & $1.043 \pm 0$ & $88.99 \pm 0.069$ & $1.04 \pm 0$ \\
 & Prox-GNN & $53.58 \pm 2.58$ & $1.04 \pm 0$ & $66.06 \pm 1.17$ & $1.04 \pm 0$ & $88.78 \pm 0.049$ & $1.04 \pm 0$ \\
\bottomrule
\end{tabular}}
\end{table*}

\begin{table}[t]
\centering
\begin{tabular}{|l|c|c|}
\hline
\textbf{Hyperparameter Changed} & \textbf{Accuracy (\%)} & \textbf{Data Fit (CE)} \\ \hline
As in main paper                    & 89.16                  & 0.24                   \\ \hline
cglsIter=4                     & 87.93                  & 0.51                   \\ \hline
cglsIter=16                    & 89.14                  & 0.18                   \\ \hline
channels=128                   & 89.10                  & 0.23                   \\ \hline
channels=32                    & 88.67                  & 0.26                   \\ \hline
layers=8                       & 88.54                  & 0.20                   \\ \hline
layers=4                       & 88.13                  & 0.22                   \\ \hline
lr=0.001                       & 89.07                  & 0.23                   \\ \hline
lr=0.01                        & 88.52                  & 0.27                   \\ \hline
solveIter=16                   & 89.14                  & 0.24                   \\ \hline
solveIter=4                    & 88.87                  & 0.28                   \\ \hline
wd=5e-05                       & 89.10                  & 0.25                   \\ \hline
wd=5e-04                       & 88.94                  & 0.22                   \\ \hline
\end{tabular}
\caption{Comparison of hyperparameter changes with accuracy and data fit (cross-entropy loss).}
\label{tab:hyperparams_sensetivity}
\end{table}

\end{document}